\documentclass[10pt,twocolumn,letterpaper]{article}

\usepackage{cvpr}              

\usepackage{graphicx}
\usepackage{amsmath}
\usepackage{amssymb}
\usepackage{booktabs}
\usepackage{graphicx}
\usepackage{siunitx}

\usepackage{multirow}
\usepackage{nth}
\usepackage{enumerate, url, pifont, textcomp, cite, xcolor}
\usepackage{bm}
\usepackage{enumitem}
\usepackage{pifont}
\usepackage[ruled,vlined]{algorithm2e}
\usepackage{color, colortbl}
\usepackage{footmisc}
\definecolor{Gray}{gray}{0.85}
\definecolor{LightCyan}{rgb}{0.88,1,1}

\newcommand\eqdef{\mathrel{\overset{\makebox[0pt]{\mbox{\normalfont\tiny\sffamily def}}}{=}}}

\usepackage[pagebackref,breaklinks,colorlinks]{hyperref}

\usepackage[capitalize]{cleveref}
\crefname{section}{Sec.}{Secs.}
\Crefname{section}{Section}{Sections}
\Crefname{table}{Table}{Tables}
\crefname{table}{Tab.}{Tabs.}

\def\cvprPaperID{2086} 
\def\confName{CVPR}
\def\confYear{2022}

\begin{document}

\title{Autofocus for Event Cameras}
\author{Shijie Lin$^{1}$, Yinqiang Zhang$^{1}$, Lei Yu$^2$, Bin Zhou$^3$, Xiaowei Luo$^4$, Jia Pan$^1$
\thanks{J. Pan is the corresponding author. This project is supported by HKSAR RGC GRF 11202119, 11207818, T42-717/20-R, HKSAR Technology Commission under the InnoHK initiative,  National Natural Science Foundation of China, Grant 61871297, and the Natural Science Foundation of Hubei Province, China, Grant 2021CFB467. {\tt$^1$\{lsj2048, zyq507, panj\}@connect.hku.hk}, {\tt$^2$ ly.wd@whu.edu.cn}, {\tt$^3$zhoubin@buaa.edu.cn}, {\tt $^4$xiaowluo@cityu.edu.hk}.
}
\\
$^1$The University of Hong Kong \, $^2$Wuhan University \, $^3$Beihang University \, $^4$City University of Hong Kong\\
 {\small Project page: \url{https://eleboss.github.io/eaf_webpage/}}
}

\maketitle

\begin{abstract}
  Focus control (FC) is crucial for cameras to capture sharp images in challenging real-world scenarios. The autofocus (AF) facilitates the FC by automatically adjusting the focus settings. However, due to the lack of effective AF methods for the recently introduced event cameras, their FC still relies on naive AF like manual focus adjustments, leading to poor adaptation in challenging real-world conditions. In particular, the inherent differences between event and frame data in terms of sensing modality, noise, temporal resolutions, \etc, bring many challenges in designing an effective AF method for event cameras. To address these challenges, we develop a novel event-based autofocus framework consisting of an event-specific focus measure called event rate (ER) and a robust search strategy called event-based golden search (EGS). To verify the performance of our method, we have collected an event-based autofocus dataset (EAD) containing well-synchronized frames, events, and focal positions in a wide variety of challenging scenes with severe lighting and motion conditions. The experiments on this dataset and additional real-world scenarios demonstrated the superiority of our method over state-of-the-art approaches in terms of efficiency and accuracy. 
\end{abstract}
\section{Introduction}
\begin{figure}[t]
  \centering
  \includegraphics[clip,width=\columnwidth]{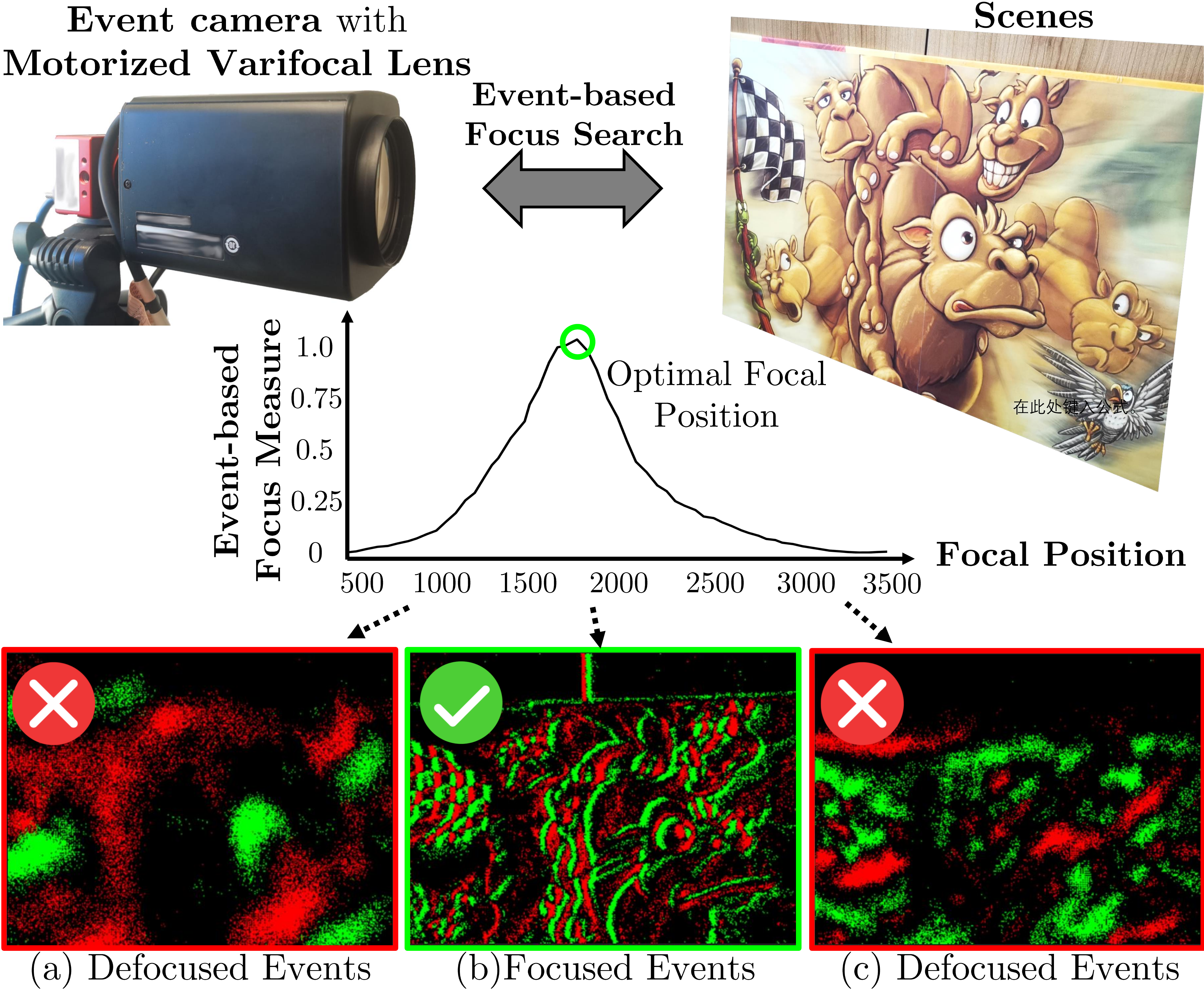}%
  \caption{Our event-based autofocus system consists of an event camera and a motorized varifocal lens. It leverages the proposed event-based focus measure and search method to focus the camera to the optimal focal position. When appropriately focused, the event camera's imaging result (b) is sharper and more informative than (a), (c) where it is defocused.}
  \label{fig:general}
  \vspace{-0.3cm}
\end{figure}
Recently, a novel neuromorphic vision sensor called event camera~\cite{DAVIS, dvs128} has gained growing attention with its significant advantages like high dynamic range (HDR, $>$\SI{130}{dB}~\cite{DAVIS}) and low latency (\SI{1}{us}~\cite{gehrig2020eklt}). Thus far, event cameras have been widely adopted in various applications, including robotics and computer vision~\cite{9138762}, where the focus control (FC) is essential for reliable perception. Similar to conventional frame-based cameras, the focused events (\cref{fig:general}b) exhibit sharper textures and convey more information than the defocused ones (\cref{fig:general}a and \cref{fig:general}c). Thus, an effective autofocus (AF) method for event cameras is in demand, especially for challenging real-world scenarios.

Conventional AF methods facilitate FC by leveraging the inherent properties of image frames like image gradient~\cite{firestone1991comparison, geusebroek2000robust, chern2001practical}, image frequency~\cite{yang2003wavelet, kautsky2002new}, and image statistics~\cite{chern2001practical}. However, events have fundamental differences with frames in terms of modality, noise, temporal resolutions, \etc, which make conventional frame-based AF methods not applicable. When developing AF methods specific for event cameras, four particular challenges are:
\begin{itemize}
   \itemsep0em
   \item \textbf{Focus measure function:} Focus measure is a critical function measuring the degree of defocus, but existing frame-based focus measures are designed for 2D image frames but not for the asynchronous event data. 
   \item \textbf{Data modality:} Unlike images, events are a stream of four-attribute tuples captured asynchronously with high temporal resolution. 
   \item \textbf{Noise:} Events contain noise that are poorly characterised~\cite{9138762} and difficult to be filtered out.
   \item \textbf{Data sizes:} During focusing, event-based AF needs to handle millions of events, making the real-time processing more challenging than conventional AF, which only needs to handle less than one hundred images. 
\end{itemize}

A naive solution to the above challenges is to rely on existing event-based image reconstruction methods~\cite{brandli2014real, munda2018real, Rebecq19pami} to convert events to images and then feed these images to conventional frame-based AF methods. However, the noise of events affects the quality of reconstructed frames and limits the performance of subsequent AF methods. In addition, as a fundamental front-end for numerous event-based applications, AF must be time-efficient, but the learning-based reconstruction is time-consuming. We handle the above challenges by developing an event-based autofocus framework from scratch. First, we developed, according to our knowledge, the first event-based focus measure leveraging the statistics of even rate (ER), which is a simple metric effective for measuring the event data captured at different focal positions. Then, we propose the event-based golden search (EGS) to cooperate with our focus measure to find the optimal focal position. EGS is invariant to the parameter of event accumulation interval and thus can robustly operate in challenging conditions, such as situations with extremely low lighting ($<$1 Lux) and violent camera shaking. 
In summary, our contributions are threefold: 
\begin{itemize}
   \itemsep0em
   \item We propose a novel event-based focus measure, \ie, event rate (ER), to measure the focus score of event data at different focal positions. It is efficient, easy to implement, and robust to noise. 
   \item Using our event-based focus measure, we propose a robust and efficient method to focus the event camera by solving a 1-dimensional optimization, which works excellently, especially in complex dynamic and low lighting conditions.
   \item We collected an event-based autofocus dataset (EAD), containing data under a wide variety of motion and lighting conditions. Extensive evaluations and comparisons have been conducted on EAD and real-world scenarios. 
\end{itemize}
\section{Related Work} 
\noindent{\textbf{Frame-based autofocus.}}
Conventional frame-based autofocus methods usually consist of two main components, \ie, the focus measure function and the search method.
The focus measure is a unimodal function that scores the imaging data captured at different focal positions. Previous works on focus measure could be categorized into contrast-based, transform-based, statistics-based, and miscellaneous methods. Contrast-based methods~\cite{firestone1991comparison, geusebroek2000robust, chern2001practical, subbarao1993focusing, nayar1994shape} leverage the property that focused images are sharper than defocused ones and use the gradient to evaluate the image sharpness. Transform-based methods~\cite{yang2003wavelet, kautsky2002new, xie2006wavelet, kristan2006bayes, lee2008enhanced, lee2009reduced, de2013image, tao2011time} work in the frequency domain and leverage the energy or frequency ratio to assess the image sharpness. Statistics-based methods~\cite{firestone1991comparison, chern2001practical, liu2016image, wee2007measure, yap2004image} attempt to reflect the degree of defocus using statistical characteristics of the image data. Other types of methods include leveraging image curvature \cite{helmli2001adaptive} and recently introduced learning-based methods \cite{herrmann2020learning}. However, these previous works all operate on image inputs in terms of a frame-like tensor. In contrast, the event data is a set of four-attribute spatial-temporal 
tuples with modality different from that of images. Thus, previous frame-based focus measures cannot be directly exploited to evaluate the focus score for the event camera.

The search methods~\cite{kehtarnavaz2003development, he2003modified, xiong1993depth, krotkov2012active, wang2020intelligent} rely on the focus score computed from images captured at different times to compute the optimal focal position. The global search obtains the optimal result by traversing all possible positions. Later efforts boost the search speed using better sequencing methods, including binary search~\cite{kehtarnavaz2003development}, hill-climbing search~\cite{he2003modified}, Fibonacci search~\cite{xiong1993depth, krotkov2012active}, \etc. Recent work using deep-learning can directly predict the optimal focal position with a few defocused images~\cite{wang2020intelligent}. However, these approaches were specifically designed for image inputs where each individual image provides rich information in the spatial domain with its large number of pixels though the solution in the temporal domain is low. 
In contrast, each individual event contains little information, but the set of all asynchronously collected events provides high resolution in the temporal domain though with relatively low resolution in the spatial domain. Thus, novel search methods specific to the event data are necessary for tackling the challenging spatial-temporal imbalance in event data.

\noindent{\textbf{{Event-based image reconstruction.}}
Frame-based AF algorithms can work on images reconstructed from events. Several recent works~\cite{brandli2014real, munda2018real, Rebecq19pami} demonstrate the possibility of reconstructing intensity images from events. Brandli~\etal~\cite{brandli2014real} use direct integration to reconstruct frames from events. Munda~\etal~\cite{munda2018real} propose a variational model for the event camera and generate events accordingly. But those methods heavily suffer from the noise. The convolutional neural network (CNN) shows outstanding performance~\cite{Rebecq19pami} in alleviating the noise, but the network's training is difficult to fit the short event accumulation intervals and its execution can be slow on low-cost computational devices, making it not applicable for applications like mobile robots. The contrast maximization framework \cite{gallego2019focus} adopts conventional focus measures as loss functions for maximizing the projected events, but it cannot solve the autofocus problem. Thus, efficient autofocus methods directly manipulating the event data are still in demand. 
\section{Problem Formulation}\label{sec:the_autofocus_measure}
The lens system can be modeled by the \textit{thick lens} model, 
\begin{align}
   \frac{1}{d_o} + \frac{1}{d_i} = \frac{1}{f}, \label{eq:lens}
\end{align}
where $d_o$ is the object distance, $d_i$ is the image distance, and $f$ is the focal length. In principle, the focused depth is uniquely determined by image-to-object distance $d_i+d_o$ and focal length $f$, and we sweep the lens to adjust the focal position $p(t)$ toward the optimum $p(t^*)$. As the lens sweep linearly, the image distance and the object distance are varying (with constant velocity) accordingly until \cref{eq:lens} is satisfied, \ie, focused. Let $\epsilon = p(t) - d_i$ be the distance error, defocus happens when the difference between $d_i$ and $p(t)$ is non-zero, i.e. $\epsilon \ne 0$. 

Generally, most AF systems follow three main assumptions. \textbf{(i) Scenes.} It should have enough textures with relatively high contrast and well-conditioned light environment without high-level flicking. \textbf{(ii) Motion.} The field of view should be constrained to the consistent targets during focusing, thus the camera motion cannot be very large. \textbf{(iii) Lens.} The lens obeys the \textit{thin lens} model and the motion of the lens motor is linear. During focusing procedure, the AF system continuously moves the lens following a specific searching path and uses time-synchronized data collected along this path as feedback to minimize $\epsilon$. Based on the data type, we can define two AF problems, \ie, the frame-based autofocus (FAF) and event-based autofocus (EAF). 

\noindent{\textbf{Frame-based autofocus (FAF).}} FAF uses images taken by frame-based cameras to estimate the optimal focal position:
\begin{align}
   p(t^*) = \operatorname*{argmax}_t F_\text{im}(S_\text{im}(t), t), \label{eq:frame_fm}
\end{align}
where $F_\text{im}$ is the frame-based focus measure function, $S_\text{im}$ is a function executing the frame-based intensity sampling from a space-time volume of perceived intensity field ($\Omega \times T \in \mathbb{R}^{2} \times \mathbb{R}$). The sampling results of $S_\text{im}$ during focusing is a set of images $\mathcal{I} = \{I_j\}_{j=1}^{N_f}$, where $N_f$ is the number of total images. As most frame-based cameras are of low FPS ($<$ \SI{30}{FPS}), the temporal resolution of the image set is relatively low. Thus, FAF only uses a few temporal samples, \ie images, to find the optimal focal position, making FAF heavily rely on each image's rich spatial content. However, as the spatial signals are easily degraded due to motion blur and low lighting, FAF usually fails to work well in challenging motion and lighting conditions. 

\noindent{\textbf{Event-based autofocus (EAF).}} Unlike FAF, EAF uses high temporal resolution events generated by event cameras to estimate the optimal focal position:
\begin{align}
   p(t^*) = \operatorname*{argmax}_t F_\text{ev}(S_\text{ev}(t, \Delta t), t), \label{eq:event_fm}
\end{align}
where $F_\text{ev}$ is the event-based focus measure function, $S_\text{ev}$ is the event-based intensity sampling function from the perceived intensity field. Since the event cameras only respond to the relative change of intensity, the sampling results of $S_\text{ev}$ is a set of events $\mathcal{E} = \{e_k : t_k\in[t-\frac{\Delta t}{2}, t+\frac{\Delta t}{2}]\}_{k=1}^{N_e}$, controlled by the sampling time $t$ and sampling interval $\Delta t$, and $N_e$ is the number of total events. Generally, $N_e$ is several orders of magnitude larger than $N_f$ because the event camera uses a pixel-wise asynchronous response to the intensity change in a very high temporal resolution. In particular, each event only contains little spatial information, \ie, a pixel's triggering time $t_k$ and polarity $p_k$. Thus, EAF needs to solve the AF problem by relying on abundant temporal samples while each sample only contains limited spatial information. Since the data modality of events is fundamentally different from that of images, challenges exist when developing efficient EAF algorithms to fully exploit the advantages of event cameras like HDR, low motion blur, and high-temporal resolutions in the EAF task.
\section{Methods}
In this section, we develop a complete EAF framework from scratch to fully exploit the event data for focusing. 
The system first traverses all focal positions to collect events (\cref{fig:pipeline}a). Then the proposed event-based golden search (\cref{sec:evsearch}) will use our event-based focus measure (\cref{sec:evfm}) to evaluate the collected events and compute the optimal focal position (\cref{fig:pipeline}b and \cref{fig:pipeline}c). 
\begin{figure*}[t]
  \centering
  \includegraphics[clip,width=1.0\linewidth]{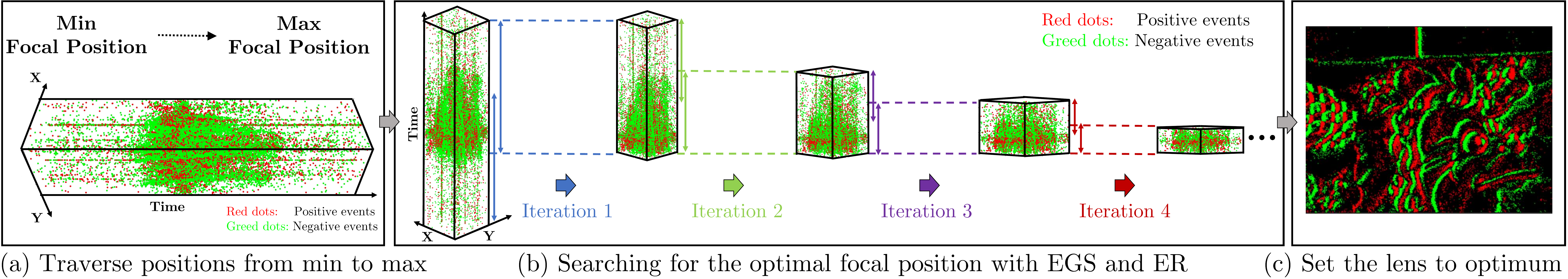}%
  \caption{Our event-based autofocus system first traverses all possible focal positions from minimum to maximum to collect events data. (b) Then the system will use the event-based golden search (EGS) in cooperation with the event-based focus measure, \ie, ER, to find the optimal focal position and (c) adjust the lens accordingly. 
  }
  \label{fig:pipeline}
  \vspace{-0.3cm}
\end{figure*}
\subsection{Preliminaries}\label{sec:preliminary}
From literature~\cite{nayar1990shape,focusing1988}, we know that defocus is a low-pass filter that attenuates high-frequency components in images. The intensity gradient represents the high-frequency component and can effectively reflect the degree of defocus~\cite{gonzalez2018digital}. Thus, conventional frame-based focus measure function in \cref{eq:frame_fm} can be defined as:
\begin{align}
  F_\text{im}(S_\text{im}(t), t) \eqdef \sum\nolimits_{\boldsymbol{x}\in \Omega} |\nabla I(\boldsymbol{x}, t)|^2,
\end{align}
where $\nabla I(\boldsymbol{x}, t) = \left(\frac{\partial I(t)}{\partial x}, \frac{\partial I(t)}{\partial y}\right)^{\top}$ is the gradient at pixel $\boldsymbol{x}$ of an intensity image sampled at time $t$. Then frame-based AF methods can solve the optimization problem in \cref{eq:frame_fm} to find the focal position. However, for the event camera,
the problem of finding an effective event-based focus measure to measure the event data is still open.
\subsection{Focus Measures for Event cameras}\label{sec:evfm}
We design an effective event-based focus measure by starting from the basic formation model of event cameras.
\subsubsection{Event Formation Model}
Event cameras~\cite{dvs128, DAVIS} asynchronously respond to the brightness change. A pixel in an event camera triggers an event $e_k$ once the change of the log intensity relative to last time step exceeds a predefined contrast threshold $C$:
\begin{align}
   \Delta L\left(\boldsymbol{x}_{k}, t_{k}\right) \eqdef L\left(\boldsymbol{x}_{k}, t_{k}\right)-L\left(\boldsymbol{x}_{k}, t_{k}-\Delta t_{k}\right)=p_{k} \cdot C, \label{eq:intensity_increament}
\end{align}
where $\boldsymbol{x}_k$ is the pixel position of the $k$-th event, $t_k$ is the triggering time, $p_{k}\in\{-1, +1\}$ is the polarity, $\Delta t_{k}$ is the duration since the last triggered event at pixel $\boldsymbol{x}_k$, and $L(\boldsymbol{x}, t) \eqdef \log I(\boldsymbol{x}, t)$ denotes the log intensity. 

The accumulation of events in a duration of $t$ can be computed by the addition of the noise and the difference between log intensities:
\begin{align}
   \label{eq:event_integration}
   \int_{0}^{t} \sum_{k} e_{k}(\boldsymbol{x}, \tau) d \tau=L(\boldsymbol{x}, t)-L(\boldsymbol{x}, 0) + \int_{0}^{t} \eta(\boldsymbol{x}, \tau) d \tau, 
\end{align}
where $ \eta(\boldsymbol{x}, t)$ is the sensor noise, which is generally unknown~\cite{9138762} and poorly characterised, and $e_{k}(\boldsymbol{x}, t)$ is the model of each event using Dirac-delta functions $\delta$~\cite{mueggler2017event}:
\begin{align}
   \label{eq:event_model}
   e_{k}(\boldsymbol{x}, t) \eqdef p_k \cdot C \cdot \delta\left(t-t_{k}\right) \delta\left(\boldsymbol{x}-\boldsymbol{x}_{k}\right).
\end{align}

\subsubsection{Event-based Focus Measure: Event Rates (ER)}
Event-based focus measures aim to efficiently provide a focus score for the events captured at different focal positions. However, classical methods for computing the gradient image~\cite{gonzalez2018digital} need spatial convolution and cannot be directly applied for events. Although we can reconstruct images from events, conducting spatial convolutions on so many reconstructed images are time-consuming. As a solution to this difficulty, \cite{scheerlinck2019asynchronous} renders gradient images by applying a linear event convolution on both sides of \cref{eq:event_integration} to skip the step of image reconstruction and directly render gradient images.  However, as discussed in \cite{scheerlinck2019asynchronous}, computing the direct integral using scheme like~\cite{munda2018real} will result in drift and biased estimations, as shown in \cref{fig:gradient_problem}}. Although the drifting can be improved by the linear filter \cite{scheerlinck2019asynchronous}, frame-wise summations are still needed to convert gradient images into focus scores and this procedure is time-consuming because the number of events can easily reach millions during focusing.

\begin{figure}[t]
  \centering
  \begin{subfigure}{0.485\linewidth}
    \includegraphics[width=\columnwidth]{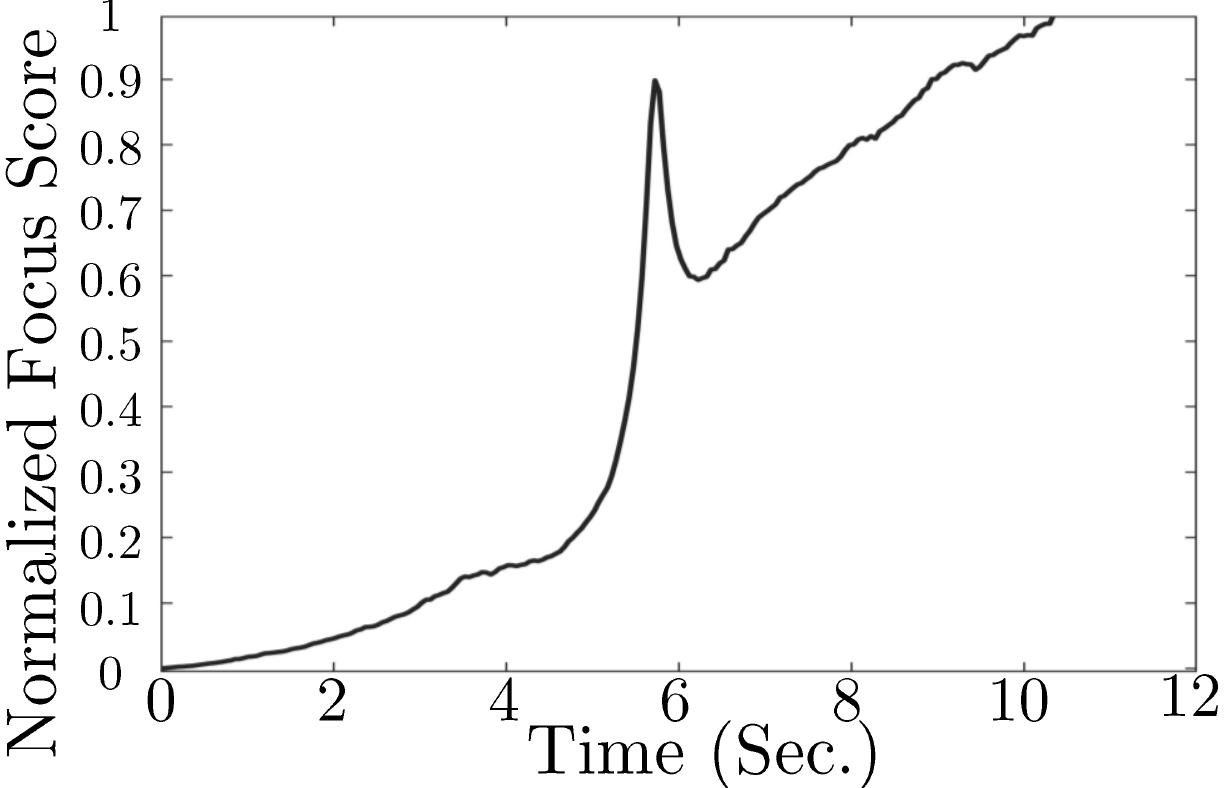}%
    \caption{Biased focus scores}
    \label{fig:gradient_problem}
  \end{subfigure}
  \hfill
  \begin{subfigure}{0.485\linewidth}
    \includegraphics[width=\columnwidth]{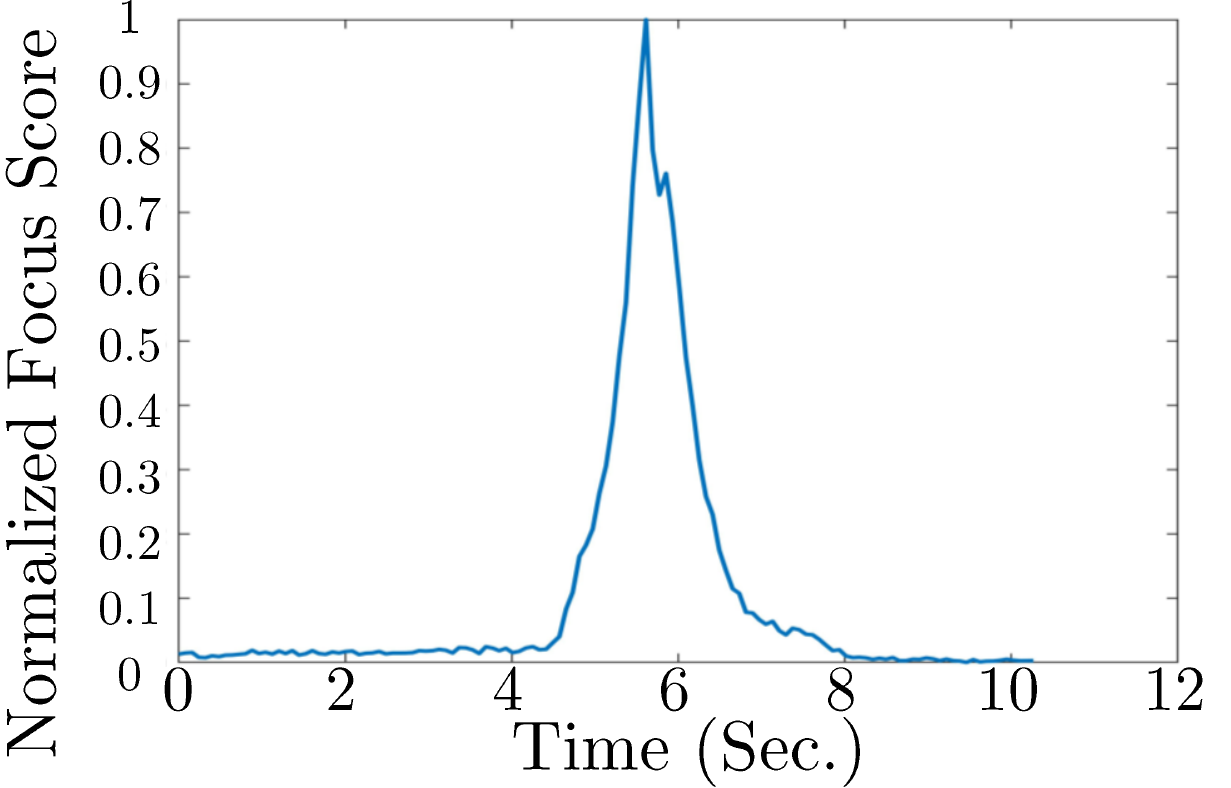}%
    \caption{Proposed ER}
    \label{fig:gradient_rate}
  \end{subfigure}
   \caption{Comparison of focus scores on sequence \textit{focus\_board\_light\_static}. (a) The focus scores using gradient computed by direct integral are affected by the noise accumulated over time, while (b) the focus scores using the event rate is immune from the accumulated noise.}
   \label{fig:gradient}
   \vspace{-0.3cm}
\end{figure}

To address the above issues, we proposed to use event rate (ER) as an event-based focus measure. First, since ER measures the rate of change, the noise will only accumulate in a short interval, largely alleviating the problem of accumulated noise, as shown in \cref{fig:gradient_rate}. Second, since the event data are transmitted in an addressed event list from the event camera to computational devices, computing ER can be simplified to count the number of events in the list without generating any frame-like array. In this way, no frame-like rendering or event-wise convolution is needed, leading to a highly efficient ER computation.

We now explain why ER could be an effective event-based focus measure. ER is computed as the average number of events within the event accumulation interval with the length $\Delta t$:
\begin{align}
  R_{e}(\boldsymbol{x}, t, \Delta t) = \frac{\int_{t-\Delta t/2}^{t+\Delta t/2} \sum_{k} \delta (t-t_{k}) \delta (\boldsymbol{x}-\boldsymbol{x}_{k}) d\tau}{\Delta t}. \label{eq:er_full}
\end{align}


As the intensity increments (\cref{eq:intensity_increament}) are caused by the moving edge~\cite{gehrig2020eklt}, given a constant velocity during the interval, the intensity changing rate can be approximated as:
\begin{align}
  \frac{\Delta L(\boldsymbol{x}, t)}{\Delta t} \approx - \nabla L(\boldsymbol{x}, t)^T \cdot \boldsymbol{v}(\boldsymbol{x}), \label{eq:gradient_speed_relation}
\end{align}
where $\frac{\Delta L(\boldsymbol{x}, t)}{\Delta t}$ denotes the intensity changing rates, $\nabla L(\boldsymbol{x}, t)$ is gradient of log intensity \cite{gehrig2020eklt}, and $\boldsymbol{v}(\boldsymbol{x})$ represents the constant velocity of pixel $\boldsymbol{x}$. By changing the interval of integration from $[0,t]$ to $[t-\Delta t/2, t +\Delta t/2]$ in \cref{eq:event_integration}, we further have:
\begin{align}
\frac{\Delta L(\boldsymbol{x}, t)}{\Delta t} = \frac{L(\boldsymbol{x}, t+\Delta t/2) - L(\boldsymbol{x}, t - \Delta t/2)}{\Delta t} \nonumber\\
    = \frac{\int_{t-\Delta t/2}^{t+\Delta t/2} \sum_{k} e_{k}(\boldsymbol{x}, \tau) d \tau - \int_{t-\Delta t/2}^{t+\Delta t/2} \eta(\boldsymbol{x}, \tau) d \tau}{\Delta t}. \label{eq:intensity_rate_2}
\end{align}
Omitting the noise term in \cref{eq:intensity_rate_2}, and replacing $e_{k}(\boldsymbol{x}, \tau)$ by the event model in \cref{eq:event_model}, we can see that the absolute value of intensity changing rate can be approximated by the proposed ER times the contrast value:
\begin{align}
  \left|\frac{\Delta L(\boldsymbol{x}, t)}{\Delta t}\right| \approx& \frac{C\int_{t-\Delta t/2}^{t+\Delta t/2} \sum_{k} \delta (t-t_{k}) \delta (\boldsymbol{x}-\boldsymbol{x}_{k}) d\tau}{\Delta t} \nonumber \\
  \label{eq:relation_gradient_er}
  =& C \cdot R_{e}(\boldsymbol{x}, t, \Delta t).
\end{align}
Combining \cref{eq:relation_gradient_er} and \cref{eq:gradient_speed_relation}, ER and gradient value are related as follows:
\begin{align}
  \label{eq:relation_er_final}
  R_{e}(\boldsymbol{x}, t, \Delta t) \approx \frac{|\nabla L(\boldsymbol{x}, t)| \cdot \boldsymbol{v}(\boldsymbol{x})}{C}.
\end{align}
Assuming a constant velocity $\boldsymbol{v}(\boldsymbol{x})$ in \cref{eq:relation_er_final}, $R_{e}(\boldsymbol{x}, t)$ is proportional to the norm of intensity gradients $\nabla L(\boldsymbol{x}, t)$, which implies that ER can be an effective indicator to the norm of intensity gradient. As mentioned in \cref{sec:preliminary}, the focal position is identified by the largest norm of intensity gradients. Then due to the proportional relationship between ER and gradient norm, the focusing location can also be found by the position with the largest ER. Therefore, ER can be an effective event-based focus measure and the focus scores is computed as:
\begin{align}
   \label{eq:fm_er}
   F_\text{ev}(S_\text{ev}(t, \Delta t), t) \eqdef \sum\nolimits_{\boldsymbol{x} \in \Omega} R^2_{e}(\boldsymbol{x}, t, \Delta t).
\end{align}
\subsection{Optimization}\label{sec:evsearch}
\begin{algorithm}[t] 
  \caption{Naive EAF with a fixed $\Delta t$}
  \KwData{accumulation interval time $\Delta t$, duration $T$\label{alg:igs}
  }
  \KwResult{Optimal focal position $p(t^*)$}
     $f_\textit{max} = 0$\;
    \For{$t\gets0$ \KwTo $T$}{
         \If{$F_\text{ev}(S_\text{ev}(t, \Delta t), t) > f_\textit{max}$}
         {
            $f_\textit{max}$ = $F_\text{ev}(S_\text{ev}(t, \Delta t), t)$\;
            $t^* = t$;
         }
    }
    
\end{algorithm}
Given the event-based focus measure in \cref{eq:fm_er}, we can solve the optimization problem in \cref{eq:event_fm} to find the optimal focal position.
During optimization, we need to tune the event accumulation interval $\Delta t$ when computing ER. We first describe a naive optimization with a manually chosen $\Delta t$ and then introduce our EAF method that can automatically choose the best $\Delta t$ and eventually the optimal $t^*$.
\subsubsection{Naive EAF with Manually Chosen $\Delta t$}\label{sec:manually_chosen}
According to our ER formulation in \cref{eq:er_full}, given a specific $\Delta t$, we can use \cref{eq:fm_er} to compute focus scores of different $t$. Among all results, we treat the $t$ that gives the maximum focus score as the optimal $t^*$. The overall algorithm is summarized in \cref{alg:igs}. An appropriate $\Delta t$ is important for this naive solution: a too large $\Delta t$ will decrease the accuracy of the finally estimated $t^*$ while a too small $\Delta t$ will generate large spiking noise in ER.
\subsubsection{EAF with Automated $\Delta t$ Search}
\begin{algorithm}[t]
  \caption{Event-based Golden Search (EGS)}
  \KwData{threshold $\mu$, golden ratio $\varphi$
  \label{alg:egs}
  }
  \KwResult{Optimal focal position $p(t^*)$}
  $[T_1, T_2] \leftarrow \text{time range of total events};$ $T =T_2 - T_1$\;  
  \While{$T > \mu$}{
     $t_1 = T_1+\frac{\varphi T}{2}$;
     $t_2 = T_2-\frac{\varphi T}{2}$;
     $\Delta t = \varphi \cdot T$\;
     \If{$F_\text{ev}(S_\text{ev}(t_1, \Delta t), t_1) \geq F_\text{ev}(S_\text{ev}(t_2, \Delta t), t_2)$}
     {
        $T_2 = \varphi \cdot T$;
     }
     \Else
     {
        $T_1 = (1-\varphi) \cdot T$;
     }
  $T\leftarrow T_2 - T_1$\;
  $t^*= (T_2 + T_1)/2$;
  }
\end{algorithm}

To make EAF immune from the artifacts of a fixed $\Delta t$, we develop the event-based golden search (EGS). EGS can automatically adjust the interval $\Delta t$ while finding $t^*$. 
The overall EGS algorithm is shown in \cref{alg:egs}. 
Specifically, EGS leverages the golden ratio $\varphi$ to divide the collected event set into two overlapping intervals and computes the focus score of each interval using \cref{eq:fm_er}. Then the entire procedure is repeated for the active interval with a higher ER. Recursively, the active interval will shrink by the same constant proportion in each step, ultimately leading to an efficient way to progressively reduce the interval locating the optimal $t^*$ with the accuracy bound $\mu$, as demonstrated in \cref{fig:pipeline}b.

Note that it is not necessary to perform event counting from scratch according to \cref{eq:er_full} whenever evaluating event rates, which is computationally redundant and has $\mathcal O(N_e^2)$ complexity. Instead, we conduct only once the event counting over the whole event set and record the partial sum results indexed by the event timestamps. In this way, the event rate for any interval can be efficiently computed as the difference of two partial sums indexed by the interval endpoints, resulting in $\mathcal O(N_e)$ complexity for EGS.


Even under perfectly static scenes, the lens motion can generate breathing effects, \eg, alternatives between blurriness and sharpness, resulting in brightness change and thus generating events. In this case, the light can spread out over multiple pixels but with attenuation due to the point spread function. Thus, for given light intensity, such attenuation reduces brightness contrast for active pixels, leading to a lower event rate, allowing our method to identify the optimal focal position. Therefore, our method can properly work in both static and dynamic scenes.
\section{Experiments}
\begin{figure}[t]
  \centering
  \begin{subfigure}{0.485\linewidth}
    \includegraphics[width=\columnwidth]{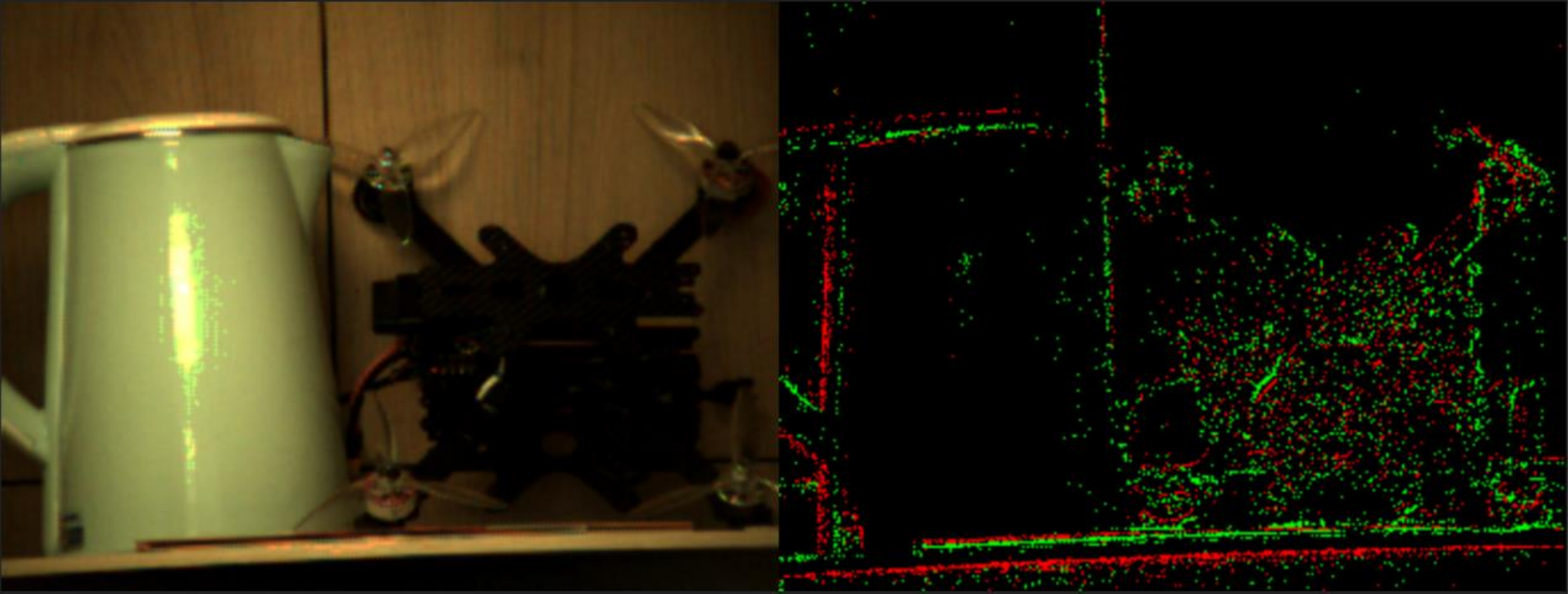}%
    \caption{\textit{bottle\_drone\_light\_static}}
    \label{fig:dataset_image_5}
  \end{subfigure}
  \hfill  
  \begin{subfigure}{0.485\linewidth}
    \includegraphics[width=\columnwidth]{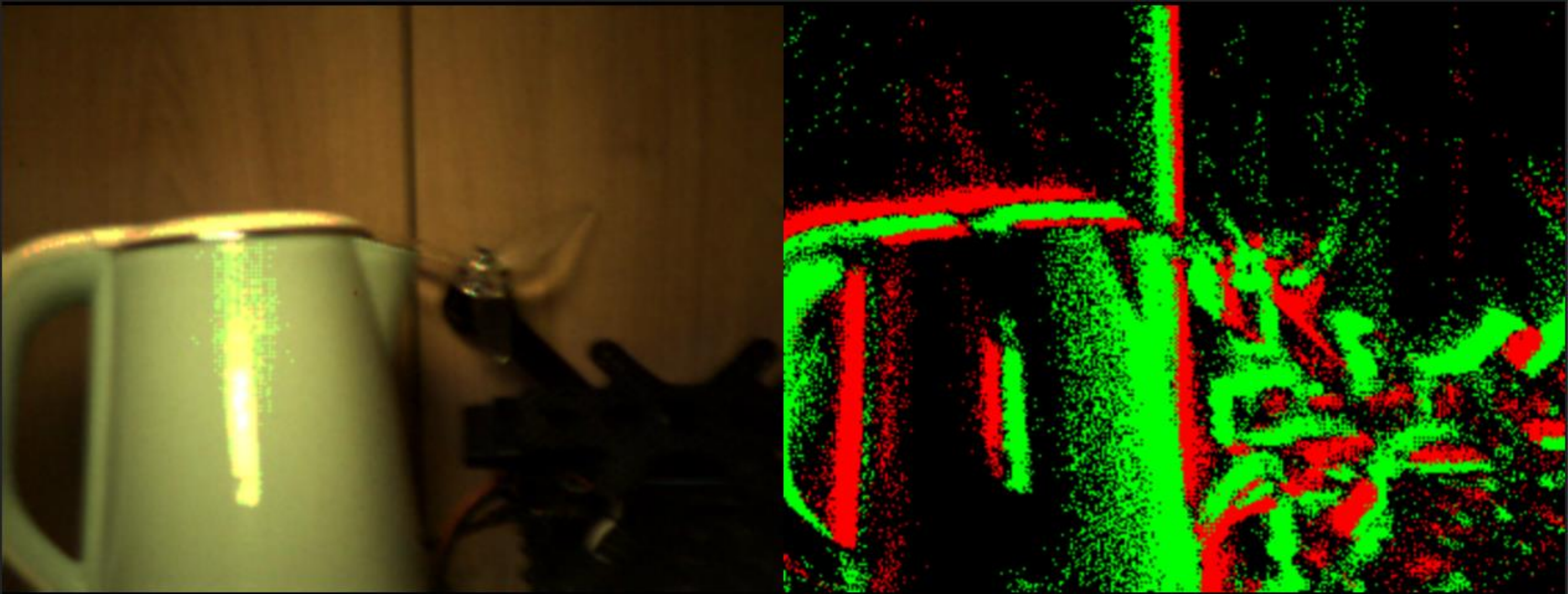}%
    \caption{\textit{bottle\_drone\_light\_dynamic}}
    \label{fig:dataset_image_6}
  \end{subfigure}
  \hfill 
  \begin{subfigure}{0.485\linewidth}
    \includegraphics[width=\columnwidth]{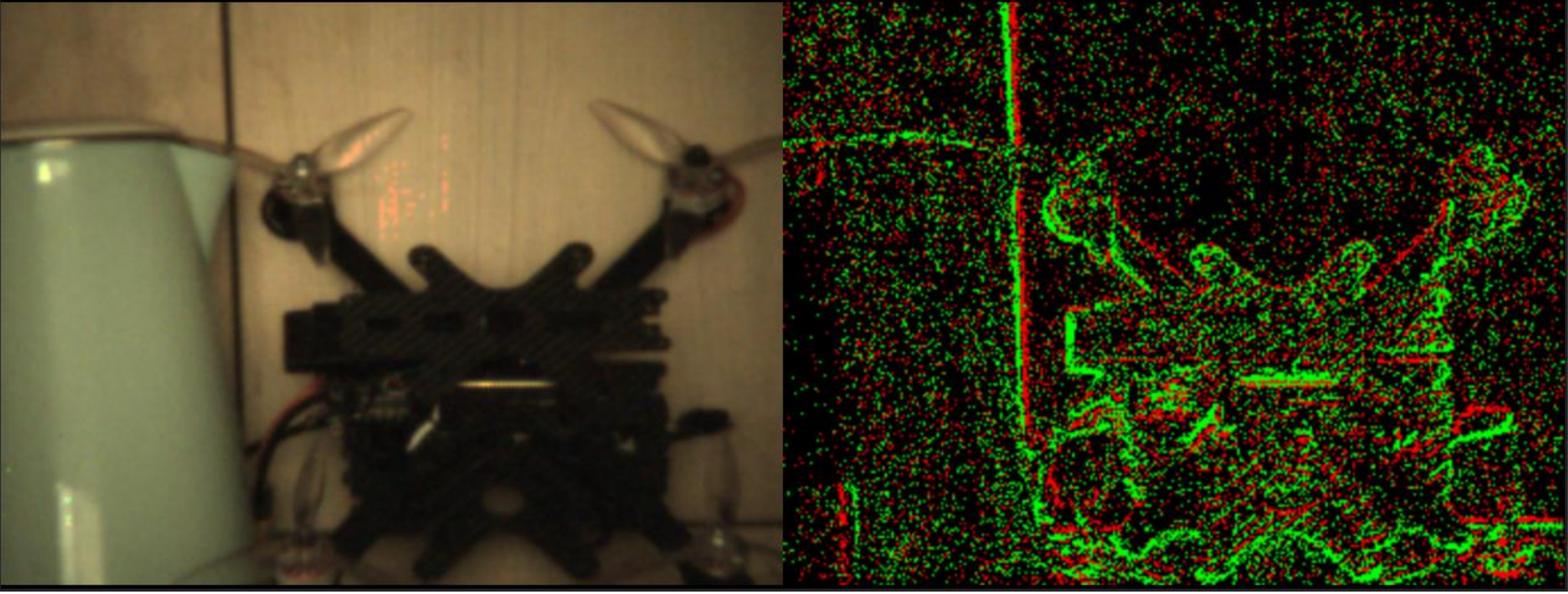}%
    \caption{\textit{bottle\_drone\_dark\_static}}
    \label{fig:dataset_image_7}
  \end{subfigure}
  \hfill
  \begin{subfigure}{0.485\linewidth}
    \includegraphics[width=\columnwidth]{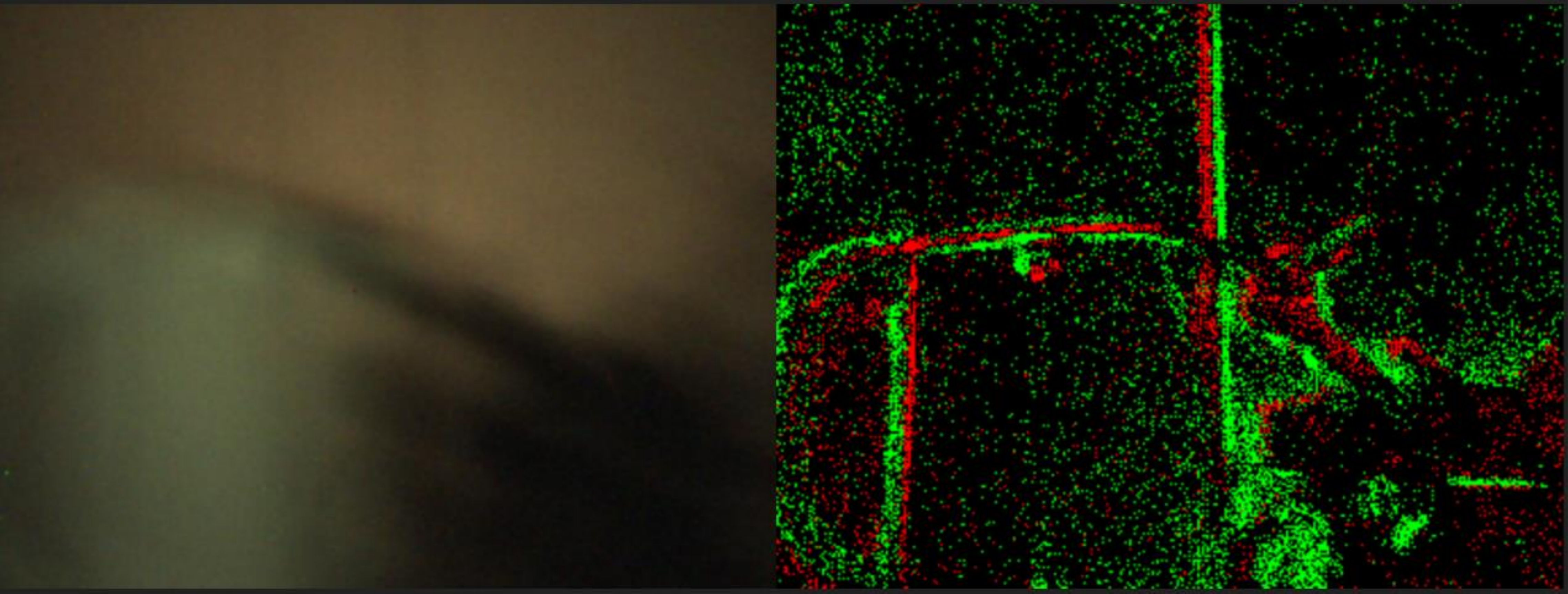}%
    \caption{\textit{bottle\_drone\_dark\_dynamic}}
    \label{fig:dataset_image_8}
  \end{subfigure}
   \caption{Examples of the event-based autofocus dataset. Our dataset contains well-synchronized frames and events in various scenes and conditions. (a), (b), (c), and (d) are captured in \textit{bottle\_drone} scene. Please see Appendix for all images in the dataset.}
   \label{fig:EAD}
   \vspace{-0.3cm}
\end{figure}
\begin{figure*}[t]
  \centering
  \begin{subfigure}{0.485\linewidth}
    \includegraphics[width=\columnwidth]{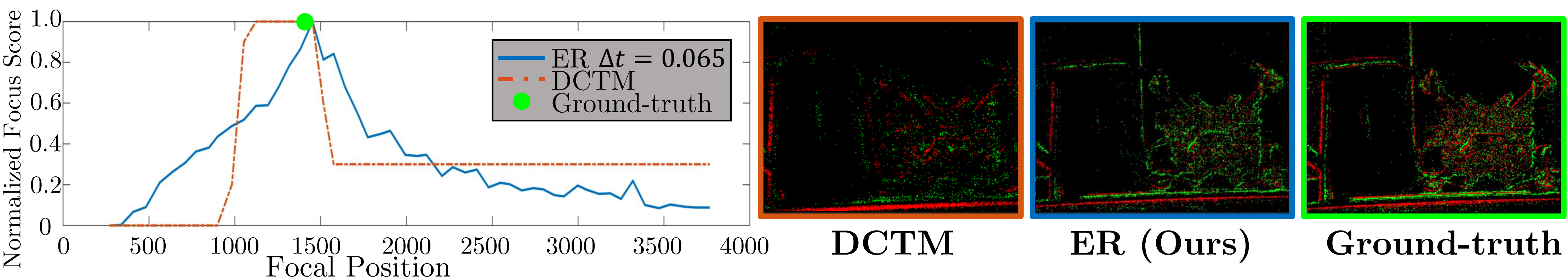}%
    \caption{Focus cruve of sequence: \textit{bottle\_drone\_light\_static}}
    \label{fig:fm_cruve_1}
  \end{subfigure}
  \hfill
  \begin{subfigure}{0.485\linewidth}
    \includegraphics[width=\columnwidth]{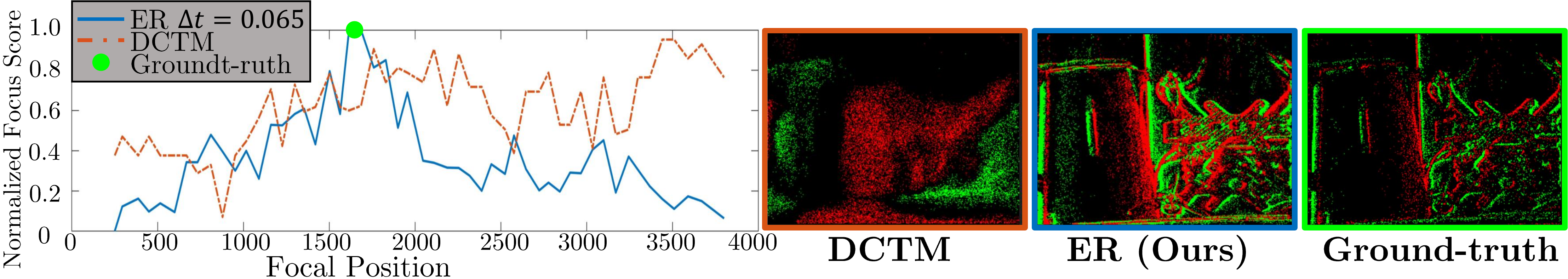}%
    \caption{Focus cruve of sequence: \textit{bottle\_drone\_light\_dynamic}}
    \label{fig:fm_cruve_2}
  \end{subfigure}
  \vfill
  \begin{subfigure}{0.485\linewidth}
    \includegraphics[width=\columnwidth]{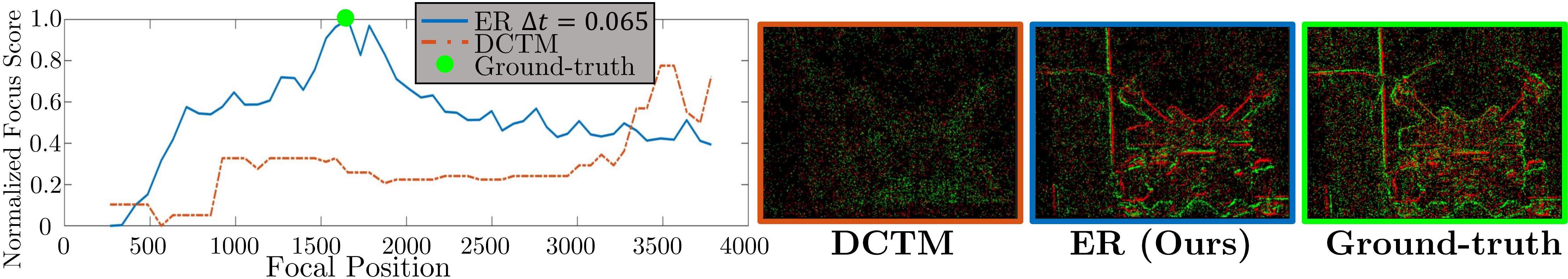}%
    \caption{Focus cruve of sequence: \textit{bottle\_drone\_dark\_static}}
    \label{fig:fm_cruve_3}
  \end{subfigure}
  \hfill
  \begin{subfigure}{0.485\linewidth}
    \includegraphics[width=\columnwidth]{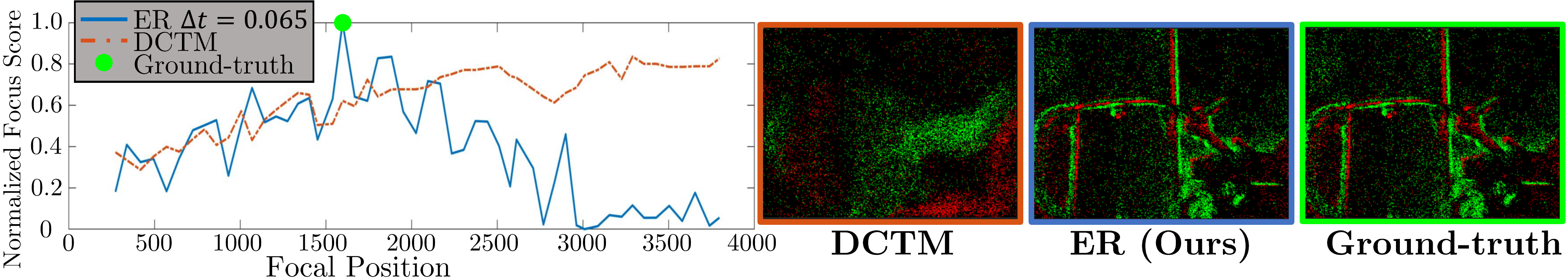}%
    \caption{Focus cruve of sequence: \textit{bottle\_drone\_dark\_dynamic}}
    \label{fig:fm_cruve_4}
  \end{subfigure}
  \caption{Focus scores in four sequences (a-d) of the proposed event-based focus measure, \ie, the event rate (ER), and the best performed frame-based focus measure in \cref{tab:compareinevents}, \ie, DCTM~\cite{lee2008enhanced}. In each sequence, the sharpest event frame is formed at the optimal focal position (the right image, green edge), in which our ER (the middle image, blue edge) gives the highest response while DCTM (the left image, orange edge) fails to do that.}
  \label{fig:fm_cruve}
  \vspace{-0.3cm}
\end{figure*}
We collect an event-based autofocus dataset (EAD) and compare our methods with several state-of-the-art baselines via a set of qualitative and quantitative experiments. We also conduct multiple tests in various real-world scenes and demonstrate our method's robustness and accuracy.
\subsection{The Event-based Autofocus Dataset (EAD)}
\noindent{\textbf{System setup.}} As briefly shown in \cref{fig:general}, we equip an event camera (DAVIS346 Color, resolution of 346$\times$260) with the motorized varifocal lens (Fujifilm D17x7.5B-YN1). We use a data acquisition card to capture the position signal of the lens and control the lens from the laptop (Intel i7-1065G7@1.3GHz). The valid focal position ranges from 220 to 3750, which linearly corresponds to the actual focal length from \SI{7.5}{mm} to \SI{140}{mm}.

\noindent{\textbf{Lighting.}} We investigate two types of lighting conditions: low lighting (refer to \textit{dark} in EAD) and normal lighting (refer to \textit{light} in EAD). The lowest lighting level can be as low as 0.7 Lux, which is challenging for conventional frame-based cameras and barely visible for human vision.

\noindent{\textbf{Motions.}} We consider two classes of motions that are common in autofocus: shaking motions (introduced by manually shaking the camera during focusing; refer to \textit{dynamic} in EAD) and without shaking motion (refer to \textit{static} in EAD). 

\noindent{\textbf{Ground-truth.}} For the ground-truth focal positions, we carefully adjust our system to capture the sharpest image of each scene and record the corresponding focal position. We then verify that this position is also the optimal one for events using high-frequency image reconstruction~\cite{EDI}.

We use the setup mentioned above to collect a dataset of frame-event sequences, with sampled snapshots shown in \cref{fig:EAD}. The dataset includes 28 sequences in four types of combination of lighting and motion conditions, i.e. \textit{static-light},  \textit{static-dark},  \textit{dynamic-light}, and  \textit{dynamic-dark}, with 7 sequences for each type. These sequences are representative for a wide variety of real-world scenes, ranging from simple indoor objects to complex outdoor construction sites. Please find the complete statistics of EAD in Appendix.
\subsection{Comparisons on Event Data Alone.}
\begin{table}[t]
   \centering
   \caption{Quantitative comparisons in events. (Better: $\downarrow$)}   \label{tab:compareinevents}
   \resizebox{\linewidth}{!}{
       \begin{tabular}{c|c|c|c|c|c|c}
           \hline
            \textbf{Method} & \textbf{Metric} & \multicolumn{5}{c}{\textbf{Scene Type}} \\
            \cline{3-7}
              &   & \multicolumn{2}{c|}{\textit{Light}} & \multicolumn{2}{c|}{\textit{Dark}} & \multirow{2}{*}{{Total}}\\
            \cline{3-6}
              &   & \multicolumn{1}{c|}{\textit{Static}} &  \multicolumn{1}{c|}{\textit{Dynamic}} & \multicolumn{1}{c|}{\textit{Static}} &  \multicolumn{1}{c|}{\textit{Dynamic}}  & \\
            \hline         
            CHEB~\cite{yap2004image} & MAE &  1842.9 & 1676.7 & 1806.3 & 1900.1 & 1806.5\\
            \cline{2-7}
            (recon. image) & RMSE &  1931.8 & 1758.3 & 1850.0 & 1947.9 & 1873.5 \\
            \hline
            HELM~\cite{helmli2001adaptive} & MAE &  1387.4 & 2001.7 & 1913.9 & 1900.3 & 1800.8 \\
            \cline{2-7}
            (recon. image) & RMSE &  1508.8 & 2067.5 & 1962.0 & 1948.1 & 1883.97 \\
            \hline
            EIGV~\cite{wee2007measure} & MAE &  1948.7 & 1143.4 & 1906.3 & 2126.0 & 1781.1\\
            \cline{2-7}
            (recon. image) & RMSE &  2040.6 & 1202.6 & 1938.4 & 2146.6 & 1869.2 \\
            \hline
            GLLV~\cite{pech2000diatom} &  MAE &  1697.1 & 1634.0 & 1625.0 & 1720.7 & 1669.2 \\
            \cline{2-7}
            (recon. image) & RMSE &  1784.0 & 1714.4 & 1726.2 & 1774.3 & 1750.0 \\
            \hline
            SML~\cite{nayar1994shape} & MAE &  1697.1 & 1634.3 & 1725.3 & 1722.7 & 1694.9 \\
            \cline{2-7}
            (recon. image) & RMSE &  1783.7 & 1714.6 & 1779.0 & 1775.7 & 1763.5\\
            \hline
            LAP3~\cite{an2008shape} & MAE &  1697.0 & 1634.3 & 1725.3 & 1722.7 & 1694.8 \\
            \cline{2-7}
            (recon. image) & RMSE &  1783.6 & 1714.6 & 1779.0 & 1775.7 & 1763.5 \\
            \hline         
            GRAD~\cite{tenenbaum1971accommodation} &  MAE &  1697.0 & 1634.3 & 1725.3 & 1722.7 & 1694.8      \\
            \cline{2-7}
            (recon. image) & RMSE &  1783.6 & 1714.6 & 1779.0 & 1775.7 & 1763.5  \\
            \hline
            WAVS~\cite{yang2003wavelet} &  MAE &  1697.0 & 1634.1 & 1725.3 & 1722.7 & 1694.8  \\
            \cline{2-7}
            (recon. image) & RMSE &  1783.6 & 1714.4 & 1779.0 & 1775.7 & 1763.4  \\
            \hline
            ACMO~\cite{shirvaikar2004optimal} &  MAE &  1501.9 & 931.3 & 1726.0 & 1828.7 & 1497.0\\
            \cline{2-7}
            (recon. image) & RMSE &  1681.0 & 1105.2 & 1779.5 & 1860.4 & 1633.6 \\
            \hline
            SFIL~\cite{minhas20093d} &  MAE &  412.9 & 1633.9 & 1725.3 & 1721.0 & 1373.3  \\
            \cline{2-7}
            (recon. image) & RMSE &  669.7 & 1714.1 & 1779.0 & 1774.5 & 1557.3 \\
            \hline
            DCTM~\cite{lee2008enhanced} &MAE &  888.3 & 1078.3 & 1632.4 & 1381.6 & 1245.1\\
            \cline{2-7}
            (recon. image) & RMSE &  1391.0 & 1313.2 & 1758.0 & 1529.2 & 1507.3\\
            \hline
            \hline  
            ER & MAE &  98.6 & 105.3 & 59.3 & 87.4 & 87.6     \\
            \cline{2-7}
            ($\Delta t=0.055$) & RMSE &  112.8 & 119.2 & 73.2 & 113.9 & 106.4   \\
            \hline  
            ER & MAE &  91.4 & 74.3 & 58.0 & 67.1 & 72.7     \\
            \cline{2-7}
            ($\Delta t=0.065$) & RMSE &  107.3 & 87.6 & 71.9 & 103.7 & 93.7   \\
            \hline  
            \multirow{2}{*}{ER + EGS} &  MAE &  \textbf{77.3} \cellcolor{Gray} & \textbf{29.0} \cellcolor{Gray} & \textbf{54.1} \cellcolor{Gray} & \textbf{64.9} \cellcolor{Gray} & \textbf{56.3} \cellcolor{Gray}\\
            \cline{2-7}
             & RMSE &  \textbf{98.9} \cellcolor{Gray} & \textbf{33.4} \cellcolor{Gray} & \textbf{71.7} \cellcolor{Gray} & \textbf{96.7} \cellcolor{Gray} & \textbf{79.7} \cellcolor{Gray}\\
            \hline       
   \end{tabular}
        }
        \vspace{-0.5cm}
\end{table}

 Since most event cameras~\cite{dvs128, chen2019live, perot2020learning} only output events without image frames, it is necessary to complete the EAF task using events alone. In this experiment, we compare our event-based autofocus methods directly working on event data with a set of frame-based autofocus methods~\cite{tenenbaum1971accommodation, nayar1994shape, shirvaikar2004optimal, yang2003wavelet, minhas20093d, an2008shape, lee2008enhanced, yap2004image} operating on 100 FPS images reconstructed from events using~\cite{brandli2014real}. For event-based autofocus methods, we investigate both the naive solution with different choices of the event accumulation interval (ER with different $\Delta t$) and EAF with automated $\Delta t$ search (ER+EGS). The comparison results are summarized in \cref{tab:compareinevents}, where the focusing error is measured by the mean absolute error (MAE) and the root mean square error (RMSE) between the estimated focal position and the ground-truth, averaged over 7 sequences in each combination of lighting and motion conditions. As we can observe, existing frame-based methods cannot handle the noise in events and all of them have much larger MAE and RMSE than event-based methods, indicating their inability to estimate the optimal focal position. In contrast, EAF methods have considerably lower error than frame-based methods, demonstrating their capability to accomplish EAF task in challenging conditions, including large scene dynamics (the \textit{dynamic} class) and low lighting conditions (the \textit{dark} class). Among them, ER+EGS gives the best result in all cases. We also provide the relative error evaluated over each sequence in the Appendix, showing similar results. 

ER+EGS's outstanding performance is due to two reasons. First, our proposed ER focus measure can always give highest score around the ground-truth focal positions, as shown in \cref{fig:fm_cruve}. Such capability to distinguish optimal focal positions is robust, working in both \textit{dynamic} sequences with violent shaking (\cref{fig:fm_cruve_2} and \cref{fig:fm_cruve_4}) and dark indoor scenes with all lights turned off (\cref{fig:fm_cruve_3} and \cref{fig:fm_cruve_4}).

\begin{figure}[t]
  \centering
  \begin{subfigure}{0.495\linewidth}
    \includegraphics[width=\columnwidth]{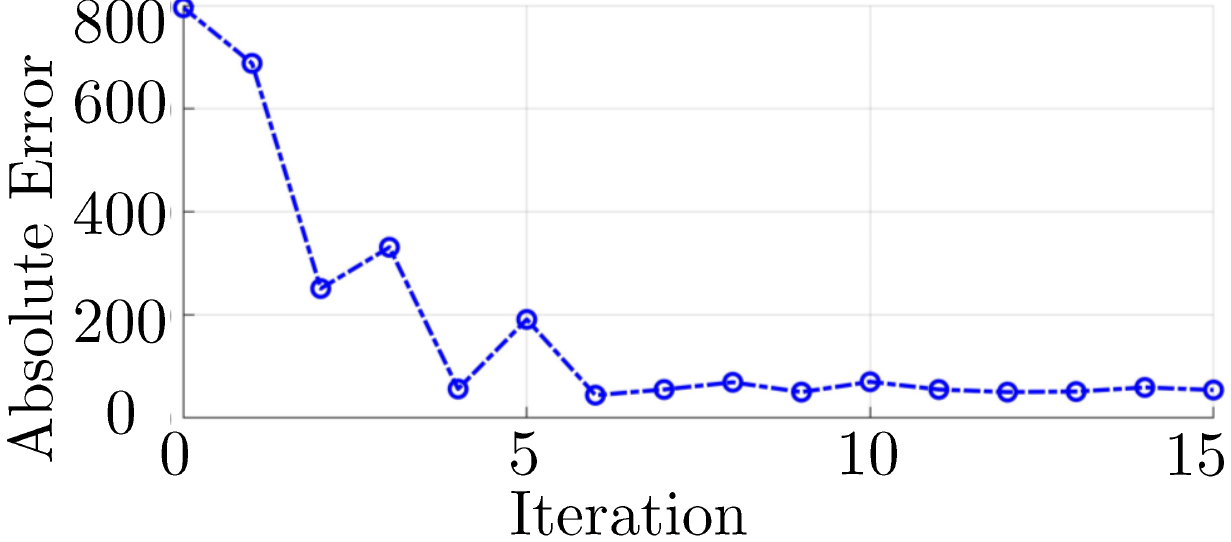}%
    \caption{\textit{bottle\_drone\_light\_static}}
    \label{fig:search_1}
  \end{subfigure}
  \hfill
  \begin{subfigure}{0.495\linewidth}
    \includegraphics[width=\columnwidth]{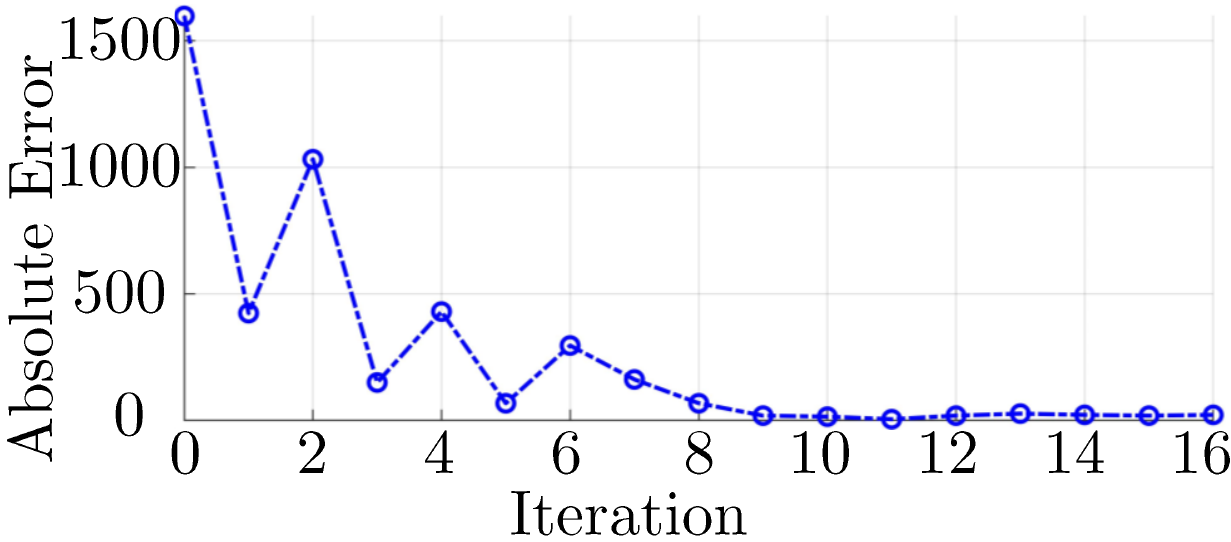}%
    \caption{\textit{bottle\_drone\_light\_dynamic}}
    \label{fig:search_2}
  \end{subfigure}
  \vfill
  \begin{subfigure}{0.495\linewidth}
    \includegraphics[width=\columnwidth]{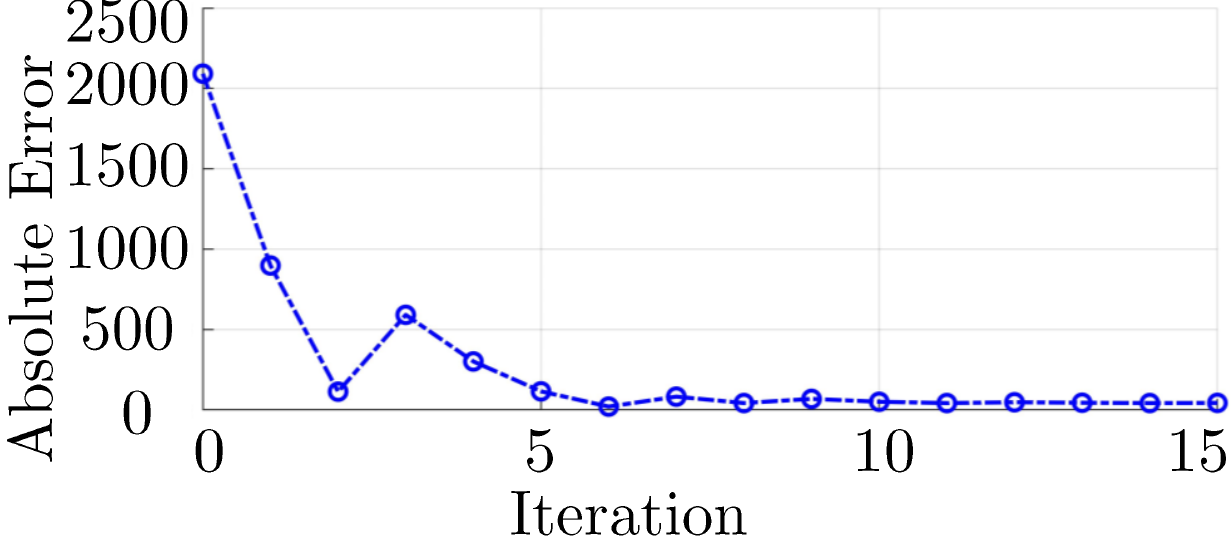}%
    \caption{\textit{bottle\_drone\_dark\_static}}
    \label{fig:search_3}
  \end{subfigure}
  \hfill
  \begin{subfigure}{0.495\linewidth}
    \includegraphics[width=\columnwidth]{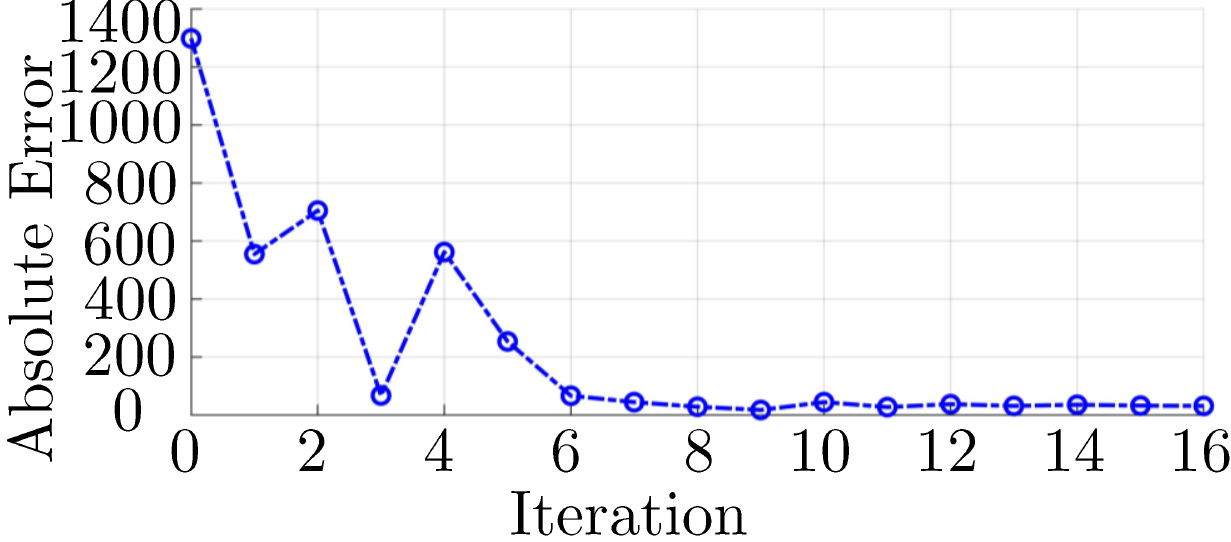}%
    \caption{\textit{bottle\_drone\_dark\_dynamic}}
    \label{fig:search_4}
  \end{subfigure}
  \caption{The curves of absolute error (AE) of focal location for the proposed event-based golden search (EGS). EGS can quickly reduce focal location error on all the four sequences in (a-d) with different lighting and motion conditions.}
  \label{fig:search}
  \vspace{-0.3cm}
\end{figure}

The second reason is EGS algorithm can quickly reduce the focal location error by adaptively tuning the event accumulation interval $\Delta t$ along with the search procedure. As shown in \cref{fig:search}, we find a fixed stopping threshold $\mu = 0.001$ works well in all challenging conditions with low lighting and violent camera shaking and the resulting ER+EGS is consistently better than the naive search with different values of fixed $\Delta t$.
\subsection{Comparisons Across Frames and Events}
\begin{table}[t]
   \centering
   \caption{Cross-comparisons in events and frames. (Better: $\downarrow$)}
   \label{tab:cross_compare_event_image}
   \resizebox{\linewidth}{!}{
       \begin{tabular}{c|c|c|c|c|c|c}
           \hline
            \textbf{Method} & \textbf{Metric} & \multicolumn{5}{c}{\textbf{Scene Type}} \\
            \cline{3-7}
            &   & \multicolumn{2}{c|}{\textit{Light}} & \multicolumn{2}{c|}{\textit{Dark}} & \multirow{2}{*}{{Total}}\\
            \cline{3-6}
              &   & \multicolumn{1}{c|}{\textit{Static}} &  \multicolumn{1}{c|}{\textit{Dynamic}} & \multicolumn{1}{c|}{\textit{Static}} &  \multicolumn{1}{c|}{\textit{Dynamic}}  & \\
            \hline         
            ACMO~\cite{shirvaikar2004optimal} &  MAE &  1431.7 & 1382.9 & 1880.9 & 1726.7 & 1605.5\\
            \cline{2-7}
            (color image) & RMSE &  1573.7 & 1517.5 & 1931.3 & 1791.1 & 1711.5\\
            \hline
            DCTM~\cite{lee2008enhanced} & MAE &  432.1 & 594.0 & 1215.1 & 1630.1 & 967.9\\
            \cline{2-7}
            (color image) & RMSE &  579.5 & 780.8 & 1448.7 & 1680.5 & 1211.2\\
            \hline
            EIGV~\cite{wee2007measure} & MAE &  690.4 & 658.1 & 951.3 & 1297.6 & 899.4\\
            \cline{2-7}
            (color image) & RMSE &  1010.7 & 936.7 & 1217.9 & 1420.4 & 1161.8 \\
            \hline
            CHEB~\cite{yap2004image} & MAE &  \textbf{37.3}\cellcolor{Gray} & 347.6 & 499.1 & 1425.6 & 577.4\\
            \cline{2-7}
            (color image) & RMSE &  \textbf{50.6}\cellcolor{Gray} & 792.7 & 781.6 & 1587.8 & 969.9 \\
            \hline
            SML~\cite{nayar1994shape} & MAE &  44.0 & 51.1 & 483.9 & 1380.0 & 489.8\\
            \cline{2-7}
            (color image) & RMSE &  52.0 & 71.7 & 787.2 & 1628.4 & 905.4\\
            \hline
            WAVS~\cite{yang2003wavelet} &  MAE &  44.0 & 56.3 & 405.1 & 1060.7 & 391.5 \\
            \cline{2-7}
            (color image) & RMSE &  52.0 & 77.2 & 768.3 & 1415.8 & 806.8 \\
            \hline
            LAP3~\cite{an2008shape} & MAE & 41.6 & 53.0 & 124.0 & 1060.9 & 319.9\\
            \cline{2-7}
            (color image) & RMSE &  48.8 & 75.7 & 165.6 & 1416.0 & 714.3 \\
            \hline
            SFIL~\cite{minhas20093d} &  MAE &  41.3 & 104.4 & 124.0 & 765.7 & 258.9 \\
            \cline{2-7}
            (color image) & RMSE &  63.7 & 147.3 & 165.6 & 1070.4 & 547.5 \\
            \hline
            GLLV~\cite{pech2000diatom} &  MAE &  46.6 & 38.9 & 191.7 & 748.6 & 256.4 \\
            \cline{2-7}
            (color image) & RMSE &  68.7 & 52.7 & 266.3 & 1085.7 & 560.6\\
            \hline
            HELM~\cite{helmli2001adaptive} & MAE &  38.4 & 47.1 & 124.0 & 772.1 & 245.4 \\
            \cline{2-7}
            (color image) & RMSE &  51.2 & 70.4 & 165.6 & 1087.5 & 551.7 \\
            \hline
            GRAD~\cite{tenenbaum1971accommodation} &  MAE &  41.1 & 57.0 & 124.0 & 421.4 & 160.9     \\
            \cline{2-7}
            (color image) & RMSE &  49.3 & 72.0 & 165.6 & 854.4 & 437.3 \\
            \hline
            \hline  
            ER & MAE &  98.6 & 105.3 & 59.3 & 87.4 & 87.6     \\
            \cline{2-7}
            ($\Delta t=0.055$) & RMSE &  112.8 & 119.2 & 73.2 & 113.9 & 106.4   \\
            \hline  
            ER & MAE &  91.4 & 74.3 & 58.0 & 67.1 & 72.7     \\
            \cline{2-7}
            ($\Delta t=0.065$) & RMSE &  107.3 & 87.6 & 71.9 & 103.7 & 93.7   \\
            \hline 
            \multirow{2}{*}{ER + EGS} &  MAE &  77.3 & \textbf{29.0}\cellcolor{Gray} & \textbf{54.1}\cellcolor{Gray} & \textbf{64.9}\cellcolor{Gray} & \textbf{56.3}\cellcolor{Gray} \\
            \cline{2-7}
             & RMSE &  98.9 & \textbf{33.4}\cellcolor{Gray} & \textbf{71.7}\cellcolor{Gray} & \textbf{96.7}\cellcolor{Gray} & \textbf{79.7}\cellcolor{Gray} \\
            \hline       
   \end{tabular}
   }
    \vspace{-0.3cm}
\end{table}

\begin{figure}[t]
   \centering
   \includegraphics[width=\columnwidth]{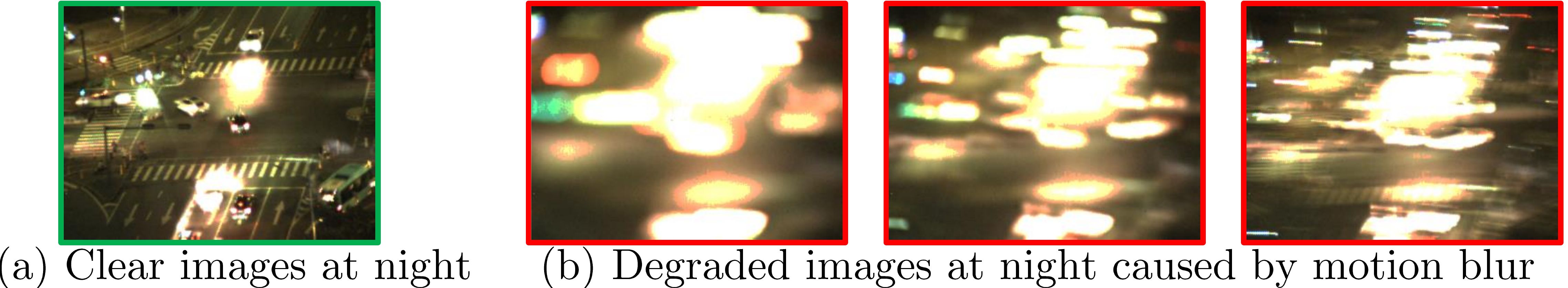}%
   \caption{(a) Clear images are easily (b) blurred by shaking.}
   \label{fig:blur}
   \vspace{-0.3cm}
\end{figure}
Because most well-established autofocus systems are designed for frame-based cameras, we also conduct cross-comparisons on both frames and events. We let conventional frame-based autofocus algorithms run on color images captured by a DAVIS346 camera during the linear focusing. Our EAF methods run on time-synchronized events captured by the same DAVIS346 camera. We report the error between the focal position with the highest focus score and the ground-truth in \cref{tab:cross_compare_event_image}. As can be observed, conventional frame-based methods work well in static scenes under proper lighting (the \textit{light} + \textit{static} class). However, as dynamics increases, the error of frame-based methods increase, especially in low light dynamic conditions (the \textit{dark} + \textit{dynamic} class) where the motions blur as shown in \cref{fig:blur} results in large errors of conventional methods. In contrast, our EAF method gives accurate estimations in all three challenging conditions, and the proposed ER+EGS achieves the lowest error except in static scenes with normal lighting. 

We also notice that some frame-based EAF methods outperform our method in relatively simple situations that are static and with good lighting conditions (\textit{static+light} column in \cref{tab:cross_compare_event_image}). The reason is a combination of noise and the simplicity of the scenes. In static scenes, events are generated by lens motions. Thus, once the scenes contain too many low-contrast textures, the activated events will be limited, which might be even reduced since the point spread function will attenuate the texture contrast. As such, the insufficient number of events will bring more bias to find the optimal focusing point since events are always contaminated by noise. Our EGS divides the total events into two intervals, and each contains an even smaller number of events, thus with lower signal-to-noise ratio (SNR), leading to sub-optimal results, increasing the ultimately averaged errors. Besides, we further provide relative error evaluated over each sequence in the Appendix. Interested readers can refer to it for detailed inspection.


\subsection{EAF in extremely low lighting conditions}
\begin{figure}[t]
  \centering
  \begin{subfigure}{\linewidth}
    \includegraphics[width=\columnwidth]{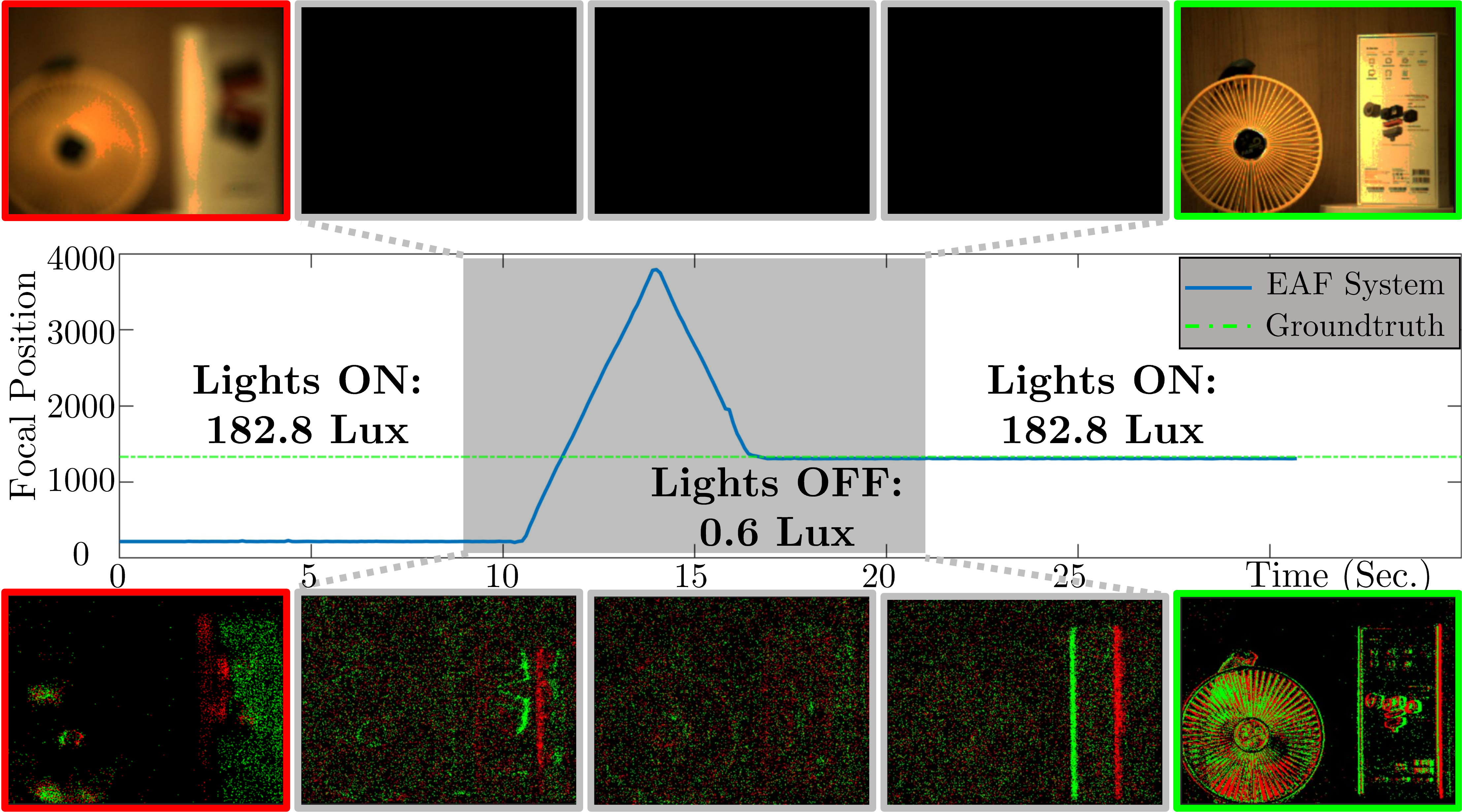}%
    \caption{Dynamic scenes: the fan is working, and the turntable is rotating the box.}
    \label{fig:extreme_1}
  \end{subfigure}
  \vfill
  \begin{subfigure}{\linewidth}
    \includegraphics[width=\columnwidth]{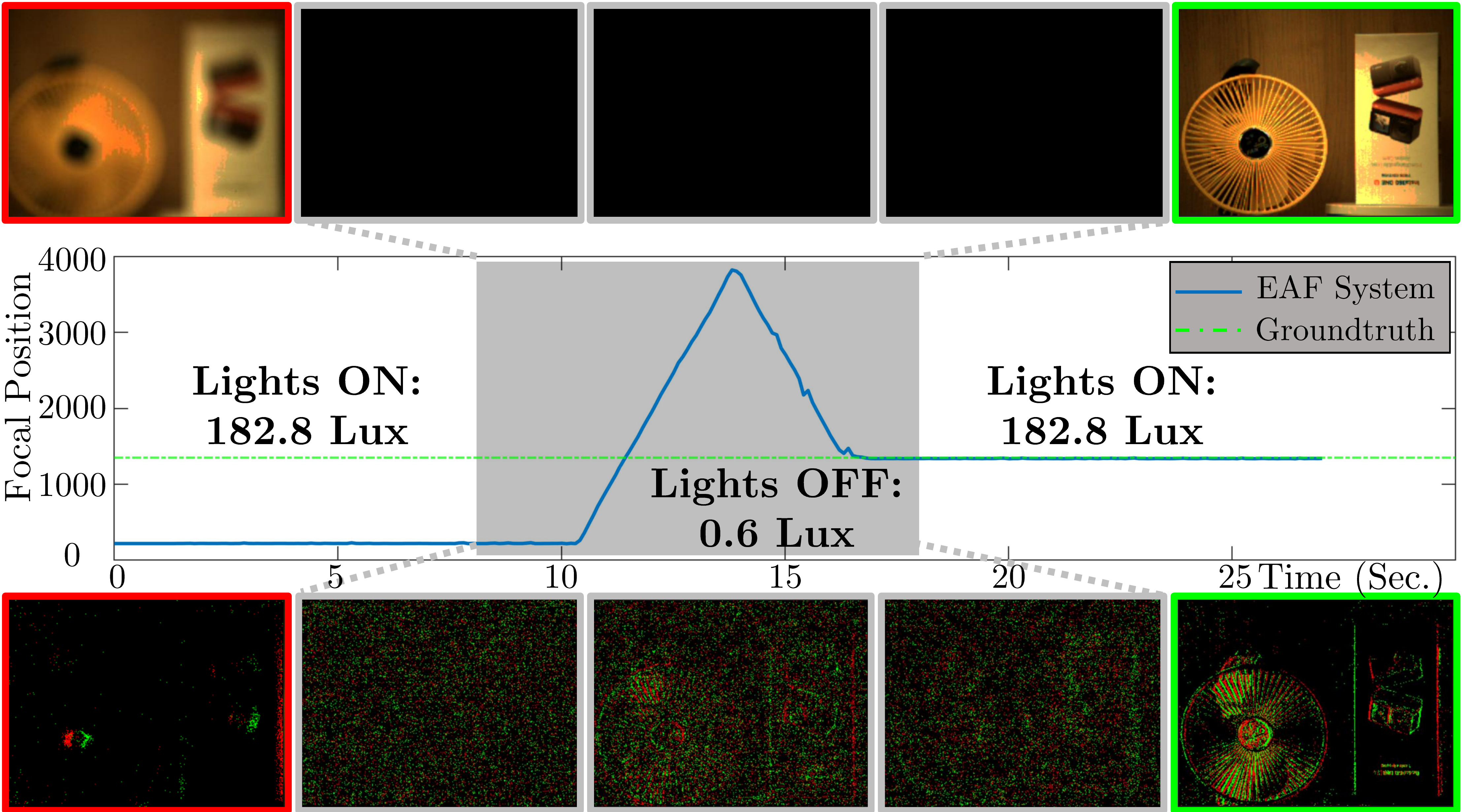}%
    \caption{Static scenes: the fan and the turntable are not working.}
    \label{fig:extreme_2}
  \end{subfigure}
  \caption{From left to right, our system accurately focus the camera in low light (marked by gray) in two situations with and without scenes motions.}
  \label{fig:extreme}
  \vspace{-0.3cm}
\end{figure}
To further illustrate the robustness of our EAF system, we perform additional AF experiments in extremely low lighting conditions, by turning the light off before the focusing procedure starts (corresponding to the gray area in \cref{fig:extreme_1} and \cref{fig:extreme_2}) in the room, and then turn the light on after the focusing finishes to check the quality of the optimal focus position estimated in dark. Results are shown in \cref{fig:extreme}, demonstrating that our EAF system can sharply focus images and events in a very dark room ($<$ \SI{0.7}{Lux}), while due to the low dynamic range of frame-based parts in the DAVIS camera, images captured in the low light are heavily degraded and cannot support the AF tasks. More details are in the video in supplemental materials.

\section{Conclusion and Discussion} 
In this paper, we provide a novel solution to the event-based autofocus task, consisting of a simple and effective focus measure called event rate (ER) and a robust event-based golden search (EGS) to efficiently find the optimal focal position for sharp imaging. We also collected the EAF dataset (EAD) with well-synchronized frames, events, and focal positions in a wide variety of motion and lighting conditions. Extensive experiments have verified the accuracy of our method when focusing an event-based camera in challenging conditions with low lighting, violent camera shaking, and complex background. 

{\noindent\textbf{Discussion}}: We found that the events usually suffer from the temporal delay under higher-contrast changes. This delay breaks the synchronization and increases the focusing error. In the future, we will exploit methods like inceptive events~\cite{baldwin2019inceptive} to address this sensing limitation.

{\small
\bibliographystyle{ieee_fullname}
\bibliography{egbib}

\begin{thebibliography}{10}\itemsep=-1pt

\bibitem{an2008shape}
Youngeun An, Gwangwon Kang, Il-Jung Kim, Hyun-Sook Chung, and Jongan Park.
\newblock Shape from focus through laplacian using 3d window.
\newblock In {\em 2008 Second International Conference on Future Generation
  Communication and Networking}, volume~2, pages 46--50, 2008.

\bibitem{baldwin2019inceptive}
R Baldwin, Mohammed Almatrafi, Jason~R Kaufman, Vijayan Asari, and Keigo
  Hirakawa.
\newblock Inceptive event time-surfaces for object classification using
  neuromorphic cameras.
\newblock In {\em International Conference on Image Analysis and Recognition
  (ICIAR)}, pages 395--403. Springer, 2019.

\bibitem{DAVIS}
Christian Brandli, Raphael Berner, Minhao Yang, Shih-Chii Liu, and Tobi
  Delbruck.
\newblock A 240$\times$180 130 db 3 $\mu$s latency global shutter
  spatiotemporal vision sensor.
\newblock {\em IEEE Journal of Solid-State Circuits}, 49(10):2333--2341, 2014.

\bibitem{brandli2014real}
Christian Brandli, Lorenz Muller, and Tobi Delbruck.
\newblock Real-time, high-speed video decompression using a frame-and
  event-based davis sensor.
\newblock In {\em 2014 IEEE International Symposium on Circuits and Systems
  (ISCAS)}, pages 686--689, 2014.

\bibitem{chen2019live}
Shoushun Chen and Menghan Guo.
\newblock Live demonstration: Celex-v: a 1m pixel multi-mode event-based
  sensor.
\newblock In {\em 2019 IEEE/CVF Conference on Computer Vision and Pattern
  Recognition Workshops (CVPRW)}, pages 1682--1683, 2019.

\bibitem{chern2001practical}
N~Ng~Kuang Chern, Poo~Aun Neow, and Marcelo~H Ang.
\newblock Practical issues in pixel-based autofocusing for machine vision.
\newblock In {\em Proceedings of IEEE International Conference on Robotics and
  Automation (ICRA)}, volume~3, pages 2791--2796, 2001.

\bibitem{de2013image}
Kanjar De and V Masilamani.
\newblock Image sharpness measure for blurred images in frequency domain.
\newblock {\em Procedia Engineering}, 64:149--158, 2013.

\bibitem{firestone1991comparison}
Lawrence Firestone, Kitty Cook, Kevin Culp, Neil Talsania, and Kendall
  Preston~Jr.
\newblock Comparison of autofocus methods for automated microscopy.
\newblock {\em Cytometry: The Journal of the International Society for
  Analytical Cytology}, 12(3):195--206, 1991.

\bibitem{9138762}
Guillermo Gallego, Tobi Delbruck, Garrick~Michael Orchard, Chiara Bartolozzi,
  Brian Taba, Andrea Censi, Stefan Leutenegger, Andrew Davison, Jorg Conradt,
  Kostas Daniilidis, and Davide Scaramuzza.
\newblock Event-based vision: A survey.
\newblock {\em IEEE Transactions on Pattern Analysis and Machine Intelligence},
  pages 1--1, 2020.

\bibitem{gallego2019focus}
Guillermo Gallego, Mathias Gehrig, and Davide Scaramuzza.
\newblock Focus is all you need: Loss functions for event-based vision.
\newblock In {\em Proceedings of the IEEE/CVF Conference on Computer Vision and
  Pattern Recognition (CVPR)}, pages 12280--12289, 2019.

\bibitem{gehrig2020eklt}
Daniel Gehrig, Henri Rebecq, Guillermo Gallego, and Davide Scaramuzza.
\newblock Eklt: Asynchronous photometric feature tracking using events and
  frames.
\newblock {\em International Journal of Computer Vision}, 128(3):601--618,
  2020.

\bibitem{geusebroek2000robust}
Jan-Mark Geusebroek, Frans Cornelissen, Arnold~WM Smeulders, and Hugo Geerts.
\newblock Robust autofocusing in microscopy.
\newblock {\em Cytometry: The Journal of the International Society for
  Analytical Cytology}, 39(1):1--9, 2000.

\bibitem{gonzalez2018digital}
Rafael~C Gonzalez and Richard~E Woods.
\newblock Digital image processing, hoboken, 2018.

\bibitem{he2003modified}
Jie He, Rongzhen Zhou, and Zhiliang Hong.
\newblock Modified fast climbing search auto-focus algorithm with adaptive step
  size searching technique for digital camera.
\newblock {\em IEEE transactions on Consumer Electronics}, 49(2):257--262,
  2003.

\bibitem{helmli2001adaptive}
Franz~Stephan Helmli and Stefan Scherer.
\newblock Adaptive shape from focus with an error estimation in light
  microscopy.
\newblock In {\em Proceedings of the 2nd International Symposium on Image and
  Signal Processing and Analysis}, pages 188--193, 2001.

\bibitem{herrmann2020learning}
Charles Herrmann, Richard~Strong Bowen, Neal Wadhwa, Rahul Garg, Qiurui He,
  Jonathan~T Barron, and Ramin Zabih.
\newblock Learning to autofocus.
\newblock In {\em Proceedings of the IEEE/CVF Conference on Computer Vision and
  Pattern Recognition (CVPR)}, pages 2230--2239, 2020.

\bibitem{kautsky2002new}
Jaroslav Kautsky, Jan Flusser, Barbara Zitova, and Stanislava
  {\v{S}}imberov{\'a}.
\newblock A new wavelet-based measure of image focus.
\newblock {\em Pattern recognition letters}, 23(14):1785--1794, 2002.

\bibitem{kehtarnavaz2003development}
Nasser Kehtarnavaz and H-J Oh.
\newblock Development and real-time implementation of a rule-based auto-focus
  algorithm.
\newblock {\em Real-Time Imaging}, 9(3):197--203, 2003.

\bibitem{kristan2006bayes}
Matej Kristan, Janez Per{\v{s}}, Matej Per{\v{s}}e, and Stanislav
  Kova{\v{c}}i{\v{c}}.
\newblock A bayes-spectral-entropy-based measure of camera focus using a
  discrete cosine transform.
\newblock {\em Pattern Recognition Letters}, 27(13):1431--1439, 2006.

\bibitem{focusing1988}
Eric Krotkov.
\newblock Focusing.
\newblock {\em International Journal of Computer Vision}, 1:223--237, 1988.

\bibitem{krotkov2012active}
Eric~P Krotkov.
\newblock {\em Active computer vision by cooperative focus and stereo}.
\newblock Springer Science \& Business Media, 2012.

\bibitem{lee2008enhanced}
Sang-Yong Lee, Yogendera Kumar, Ji-Man Cho, Sang-Won Lee, and Soo-Won Kim.
\newblock Enhanced autofocus algorithm using robust focus measure and fuzzy
  reasoning.
\newblock {\em IEEE Transactions on Circuits and Systems for Video Technology},
  18(9):1237--1246, 2008.

\bibitem{lee2009reduced}
Sang-Yong Lee, Jae-Tack Yoo, Yogendera Kumar, and Soo-Won Kim.
\newblock Reduced energy-ratio measure for robust autofocusing in digital
  camera.
\newblock {\em IEEE Signal Processing Letters}, 16(2):133--136, 2009.

\bibitem{dvs128}
Patrick Lichtsteiner, Christoph Posch, and Tobi Delbruck.
\newblock A 128$\times$128 120 db 15$\mu$s latency asynchronous temporal
  contrast vision sensor.
\newblock {\em IEEE journal of solid-state circuits}, 43(2):566--576, 2008.

\bibitem{liu2016image}
Shuxin Liu, Manhua Liu, and Zhongyuan Yang.
\newblock An image auto-focusing algorithm for industrial image measurement.
\newblock {\em EURASIP Journal on Advances in Signal Processing},
  2016(1):1--16, 2016.

\bibitem{minhas20093d}
Rashid Minhas, Abdul~A Mohammed, QM~Jonathan Wu, and Maher~A Sid-Ahmed.
\newblock 3d shape from focus and depth map computation using steerable
  filters.
\newblock In {\em International conference image analysis and recognition},
  pages 573--583. Springer, 2009.

\bibitem{mueggler2017event}
Elias Mueggler, Henri Rebecq, Guillermo Gallego, Tobi Delbruck, and Davide
  Scaramuzza.
\newblock The event-camera dataset and simulator: Event-based data for pose
  estimation, visual odometry, and slam.
\newblock {\em The International Journal of Robotics Research}, 36(2):142--149,
  2017.

\bibitem{munda2018real}
Gottfried Munda, Christian Reinbacher, and Thomas Pock.
\newblock Real-time intensity-image reconstruction for event cameras using
  manifold regularisation.
\newblock {\em International Journal of Computer Vision}, 126(12):1381--1393,
  2018.

\bibitem{nayar1990shape}
Shree~K Nayar and Yasuo Nakagawa.
\newblock Shape from focus: An effective approach for rough surfaces.
\newblock In {\em Proceedings., IEEE International Conference on Robotics and
  Automation (ICRA)}, pages 218--225, 1990.

\bibitem{nayar1994shape}
Shree~K Nayar and Yasuo Nakagawa.
\newblock Shape from focus.
\newblock {\em IEEE Transactions on Pattern analysis and machine intelligence},
  16(8):824--831, 1994.

\bibitem{EDI}
Liyuan Pan, Cedric Scheerlinck, Xin Yu, Richard Hartley, Miaomiao Liu, and
  Yuchao Dai.
\newblock Bringing a blurry frame alive at high frame-rate with an event
  camera.
\newblock In {\em 2019 IEEE/CVF Conference on Computer Vision and Pattern
  Recognition (CVPR)}, pages 6813--6822, June 2019.

\bibitem{pech2000diatom}
Jos{\'e}~Luis Pech-Pacheco, Gabriel Crist{\'o}bal, Jes{\'u}s Chamorro-Martinez,
  and Joaqu{\'\i}n Fern{\'a}ndez-Valdivia.
\newblock Diatom autofocusing in brightfield microscopy: a comparative study.
\newblock In {\em Proceedings 15th International Conference on Pattern
  Recognition (ICPR)}, volume~3, pages 314--317, 2000.

\bibitem{perot2020learning}
Etienne Perot, Pierre de Tournemire, Davide Nitti, Jonathan Masci, and Amos
  Sironi.
\newblock Learning to detect objects with a 1 megapixel event camera.
\newblock {\em Advances in Neural Information Processing Systems}, 33, 2020.

\bibitem{Rebecq19pami}
Henri Rebecq, Ren{\'e} Ranftl, Vladlen Koltun, and Davide Scaramuzza.
\newblock High speed and high dynamic range video with an event camera.
\newblock {\em IEEE transactions on pattern analysis and machine intelligence},
  2019.

\bibitem{scheerlinck2019asynchronous}
Cedric Scheerlinck, Nick Barnes, and Robert Mahony.
\newblock Asynchronous spatial image convolutions for event cameras.
\newblock {\em IEEE Robotics and Automation Letters}, 4(2):816--822, 2019.

\bibitem{shirvaikar2004optimal}
Mukul~V Shirvaikar.
\newblock An optimal measure for camera focus and exposure.
\newblock In {\em Thirty-Sixth Southeastern Symposium on System Theory, 2004.
  Proceedings of the}, pages 472--475, 2004.

\bibitem{subbarao1993focusing}
Murali Subbarao, Tae-Sun Choi, and Arman Nikzad.
\newblock Focusing techniques.
\newblock {\em Optical Engineering}, 32(11):2824--2836, 1993.

\bibitem{tao2011time}
Ran Tao, Wei Zhang, and Yanlei Li.
\newblock Time--frequency filtering-based autofocus.
\newblock {\em Signal processing}, 91(6):1401--1408, 2011.

\bibitem{tenenbaum1971accommodation}
Jay~Martin Tenenbaum.
\newblock {\em Accommodation in computer vision}.
\newblock Stanford University, 1971.

\bibitem{wang2020intelligent}
Chengyu Wang, Qian Huang, Ming Cheng, Zhan Ma, and David~J Brady.
\newblock Intelligent autofocus.
\newblock {\em arXiv preprint arXiv:2002.12389}, 2020.

\bibitem{wee2007measure}
Chong-Yaw Wee and Raveendran Paramesran.
\newblock Measure of image sharpness using eigenvalues.
\newblock {\em Information Sciences}, 177(12):2533--2552, 2007.

\bibitem{xie2006wavelet}
Hui Xie, Weibin Rong, and Lining Sun.
\newblock Wavelet-based focus measure and 3-d surface reconstruction method for
  microscopy images.
\newblock In {\em 2006 IEEE/RSJ International Conference on Intelligent Robots
  and Systems (IROS)}, pages 229--234, 2006.

\bibitem{xiong1993depth}
Yalin Xiong and Steven~A Shafer.
\newblock Depth from focusing and defocusing.
\newblock In {\em Proceedings of IEEE Conference on Computer Vision and Pattern
  Recognition (ICCV)}, pages 68--73, 1993.

\bibitem{yang2003wavelet}
Ge Yang and Bradley~J Nelson.
\newblock Wavelet-based autofocusing and unsupervised segmentation of
  microscopic images.
\newblock In {\em Proceedings of IEEE/RSJ International Conference on
  Intelligent Robots and Systems (IROS)}, volume~3, pages 2143--2148, 2003.

\bibitem{yap2004image}
Pew~Thian Yap and P Raveendran.
\newblock Image focus measure based on chebyshev moments.
\newblock {\em IEE Proceedings-Vision, Image and Signal Processing},
  151(2):128--136, 2004.

\end{thebibliography}
}
\end{document}